\documentclass[times,review, 10pt]{elsarticle}
\usepackage{color}
\usepackage{amsmath,amsfonts,amssymb,pifont}
\usepackage{booktabs}
\usepackage{graphicx}
\usepackage{bm}
\usepackage{array}
\usepackage{subfigure}
\usepackage{multirow}
\usepackage{appendix}
\usepackage[colorlinks=true, allcolors=blue]{hyperref}
\usepackage[table,xcdraw]{xcolor}
\usepackage{lineno,hyperref}
\usepackage{wrapfig}
\usepackage{comment}
\usepackage{mathtools} 

\begin{document}

\begin{frontmatter}

\title{
Semantic Change Detection of Roads and Bridges: A Fine-grained Dataset and Multimodal Frequency-driven Detector
}

\author[First]{Qingling Shu}\ead{2563489133@qq.com}
\author[First]{Sibao Chen\corref{cor1}}
\cortext[cor1]{Corresponding author.}
\ead{sbchen@ahu.edu.cn}
\author[First]{Xiao Wang}
\author[Second]{Zhihui You}
\author[First]{Wei Lu}
\author[First]{Jin~Tang}
\author[First]{Bin~Luo}
\address[First]{
	MOE Key Lab of ICSP, IMIS Lab of Anhui Province, Anhui Provincial Key Lab of Multimodal Cognitive Computation, School of Computer Science and Technology,
	Anhui University, Hefei, China
}
\address[Second]{
School of Public Safety and Emergency Management, Anhui University of Science and Technology, Hefei, China
}

\begin{abstract}
Accurate detection of  road and bridge changes is crucial for  urban planning and transportation management, yet presents unique challenges for general change detection (CD). Key difficulties arise from maintaining the continuity of roads and bridges as linear structures and disambiguating visually similar land covers (e.g., road construction vs. bare land).
Existing spatial-domain models struggle with these issues, further hindered by the lack of specialized, semantically rich datasets. To fill these gaps, we introduce the Road and Bridge Semantic Change Detection (RB-SCD) dataset. As the first benchmark to systematically target semantic change detection of roads and bridges, RB-SCD offers comprehensive fine-grained annotations for 11 semantic change categories. This enables a detailed analysis of traffic infrastructure evolution.
Building on this,
we propose a novel framework, the Multimodal Frequency-Driven Change Detector (MFDCD).
MFDCD integrates multimodal features in the frequency domain through two key components:
(1) the Dynamic Frequency Coupler (DFC), which  
leverages wavelet transform to decompose visual features, enabling it to robustly model the continuity of linear transitions; and (2)
the Textual Frequency Filter (TFF), which 
encodes semantic priors into frequency-domain graphs and applies filter banks to align them with visual features, resolving semantic ambiguities. 
Experiments demonstrate the state-of-the-art performance of MFDCD on RB-SCD and three public CD datasets. 
The code will be available at https://github.com/DaGuangDaGuang/RB-SCD.
\end{abstract}


\end{frontmatter}


\section{INTRODUCTION}
\label{sec:intro}
In the field remote sensing (RS), change detection (CD) has become an essential task for monitoring environmental, urban, and infrastructure changes~\cite{GlobalLC2}
. With the acceleration of urbanization and the continuous development of transportation infrastructure, 
accurately identifying structural and functional changes in roads and bridges helps urban authorities to assess the current state of transportation networks, forecast future demands, and make data-driven decisions about expansions, detours, and upgrades.
Moreover, road and bridge CD enables timely identification of structural deterioration, unauthorized demolitions, or unreported construction activities~\cite{Introduction2}.

However, as shown in Tab.~\ref{tab:dataset_comparation}, traditional binary change detection (BCD) datasets such as LEVIR~\cite{LEVIR}, WHU~\cite{WHU}, CDD~\cite{CDD}, SYSU~\cite{SYSU}, merely indicate whether a change has occurred through a simple binary mask (0 for no change, 1 for change). They lack   fine-grained semantic details.
While existing semantic change detection (SCD) datasets like Landsat~\cite{Landsat-SCD}, HRSCD~\cite{HRSCD}, SECOND~\cite{SECOND} mainly focus on changes in land cover.
Changes related to roads and bridges in traffic scenes are either limited in number or completely absent. Although  existing road extraction datasets~\cite{CHN6-CUG},~\cite{Massachusetts} 
mainly focus on binary pixel segmentation of roads, they cannot be used to focus on road changes.

\begin{figure*}[t!]
	\centering
	\includegraphics[width=1\textwidth]{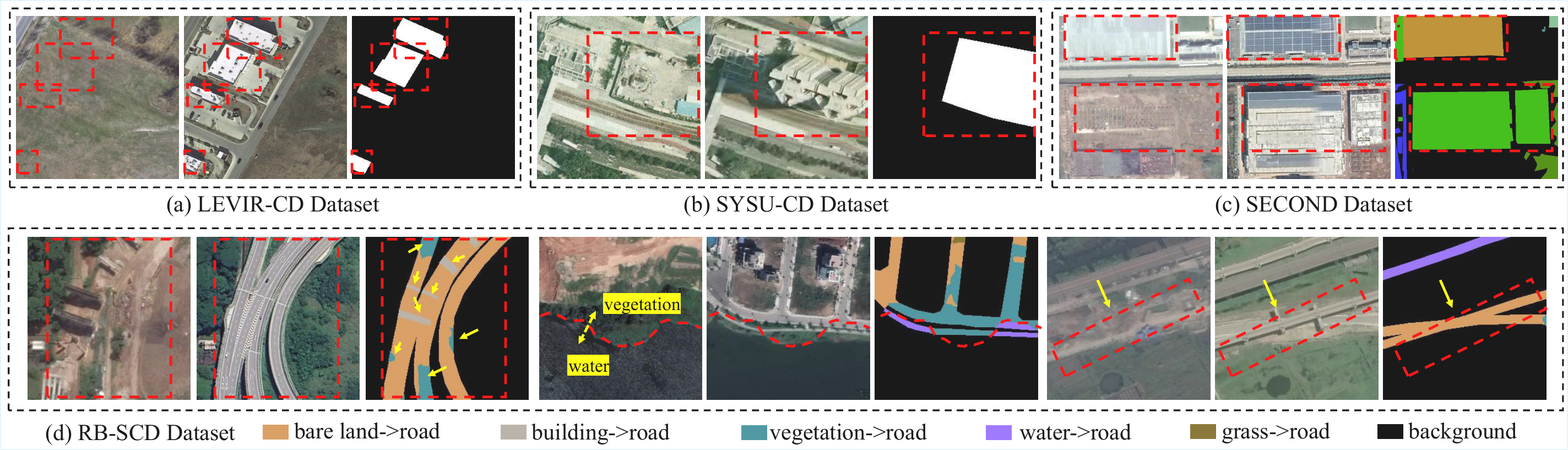} 
	\caption{Visual comparison of change characteristics across prominent CD datasets. (a) LEVIR-CD and (b) SYSU-CD primarily feature discrete, monolithic building changes,  (c) SECOND often focus on  land-cover transitions. (d) RB-SCD introduces the unique challenges of detecting linear and heterogeneous  change (source of change is heterogeneous) transitions, where winding roads are formed from multiple land-cover types. Furthermore, it presents high semantic ambiguity, with visually similar yet semantically distinct transitions such as water-\textgreater{}road and vegetation-\textgreater{}road, which are difficult for conventional models to distinguish.}
	\label{fig:datasetvis}
\end{figure*}

Methodologically, detecting semantic changes in roads and bridges presents unique challenges that are not fully addressed by general-purpose CD methods. Two key difficulties stand out. First, 
unlike the discrete, monolithic changes of buildings common in datasets (first row of Fig.~\ref{fig:datasetvis}), road networks evolve as  winding linear structures and their changes often involve complex transitions from multiple, heterogeneous land-cover types such as water, vegetation, and bare land  simultaneously (second row of Fig.~\ref{fig:datasetvis}).
\begin{table*}
	\centering
	\caption{Comparison of
		public remote sensing change detection and road extraction datasets. 
		For image counting, the unit for change detection datasets is pairs of images, whereas for road extraction datasets, it is individual image. Notably, no existing dataset systematically addresses road and bridge semantic changes.}
	\label{tab:datasets}
	\resizebox{1\textwidth}{!}{
		\begin{tabular}{lcccccc}
			\midrule
			Dataset & Resolution (m) & Image count & Image size (pixels) & Type & Objects of interest & Classes \\
			\hline
			\rowcolor{gray!20} 
			\multicolumn{7}{l}{Change detection}
			\\
			LEVIR-CD ~\cite{LEVIR} & 0.5 & 637 & $1024 \times 1024$ & Binary & Buildings & -- \\
			WHU  ~\cite{WHU} & 0.2 & 1 & $32207 \times 15354$ & Binary & Buildings & -- \\
			CDD ~\cite{CDD} & 0.03 - 1 & 16000 & $256 \times 256$ & Binary & Buildings & -- \\
			SYSU ~\cite{SYSU} & 0.5 & 20000 & $256 \times 256$ & Binary & Buildings & -- \\
			Landsat-SCD ~\cite{Landsat-SCD} & 30 & 8468 & $416 \times 416$ & Semantic& Land cover & 4 \\
			HRSCD ~\cite{HRSCD} & 0.5 & 291 & $10000 \times 10000$ & Semantic & Land cover & 5 \\
			SECOND ~\cite{SECOND} & 0.5 - 3 & 4662 & $512 \times 512$ & Semantic & Land cover & 6 \\
			Hi-UCD ~\cite{Tian2020HiUCD} & 0.1 & 1293 & $1024 \times 1024$ & Semantic & Land cover & 9 \\
			\hline
			\rowcolor{gray!20}
			\multicolumn{7}{l}{Road extraction} \\  
			CHN6-CUG ~\cite{CHN6-CUG} & 0.5 & 4511 & $512 \times 512$ & Binary & Road & -- \\
			Massachusetts ~\cite{Massachusetts} & 1.5 & 1000 & $1500 \times 1500$ & Binary & Road & -- \\
			\midrule
			
			RB-SCD (Ours) & 0.59 & 260 & $270 \times 534$ - $7215 \times 4366$ & Semantic & Roads and bridges & 8 \\
			\bottomrule
	\end{tabular}}
	\label{tab:dataset_comparation}
\end{table*}
The limited receptive fields of recent methods like SUNNet~\cite{SNUNet}, ICIF~\cite{ICIF}, USSFCNet~\cite{USSFC-Net}  prevent them from perceiving the overall structure of road networks. Consequently, while they may correctly identify numerous localized patches of change, they ultimately produce fragmented and disconnected change maps, failing to assemble these patches into a single, continuous linear feature.
Standard attention  models such as BIT~\cite{BIT}, Changeformer~\cite{Changeformer}, DMINet~\cite{DMINet}, ELGCNet~\cite{ELGC-Net} 
leverage global attention to overcome this locality, their standard mechanisms treat all spatial relationships equally, lacking the explicit inductive bias required to model heterogeneous, linear structures.
Second, the task is plagued by high semantic ambiguity. Visual features alone are often insufficient to resolve ambiguities, such as distinguishing a road under construction from bare land (the first and third sample of Fig.~\ref{fig:datasetvis} (d)), or differentiating shallow water from adjacent vegetation (second sample of Fig.~\ref{fig:datasetvis} (d)).
Even multimodal  models like ChangeCLIP~\cite{ChangeCLIP}, which leverage vision-language for CD, may struggle when visual features themselves are highly ambiguous, as their fusion process operates within the noisy spatial domain.


\begin{figure*}[t!]
	\centering
	\includegraphics[width=1\textwidth]{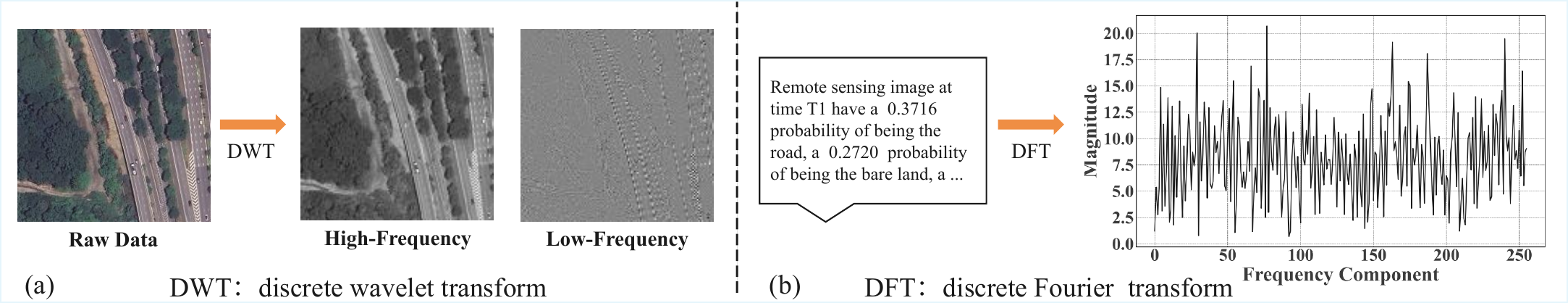} 
	\caption{
		Frequency-domain transformation of visual and textual data. (a) The DWT separates a visual scene into high-frequency (details) and low-frequency (structure) components. (b) The DFT converts a textual description into a global frequency spectrum, serving as a robust semantic signature for  fusion.
	}
	\label{fig:Frequency_analysis}
\end{figure*}

The limitations of existing spatial-domain methods highlight the need for a new approach that operates in a more suitable domain. To robustly model linear structures, the core insight is that their essential continuity of structure is naturally encoded in low-frequency components, while sharp transition boundaries correspond to high-frequency components. This intrinsic property makes the frequency domain a logical choice. As shown in Fig.~\ref{fig:Frequency_analysis}(a), we leverage the discrete wavelet transform (DWT) to decompose visual features and explicitly isolate these essential structural forms from fine-grained boundary details.
To resolve deep semantic ambiguity, multimodal integration is required. However, prior methods suffer from fusing textual descriptions directly with unstable pixel-level features in the spatial domain. 
We instead transform text into its frequency spectrum via discrete Fourier transform (DFT) (Fig.~\ref{fig:Frequency_analysis}(b)).
This frequency-domain representation naturally filters out minor textual variations, providing a stable basis for a cleaner alignment with visual features and overcoming the core limitations of spatial-level fusion.

Based on these insights, 
we first introduce the \textbf{R}oad and \textbf{B}ridge \textbf{S}emantic \textbf{C}hange \textbf{D}etection (\textbf{RB-SCD}) dataset, a new benchmark specifically designed for SCD of roads and bridges. RB-SCD provides fine-grained semantic annotations for 11 distinct change categories across diverse geographic regions, enabling a detail evaluation of model capabilities in traffic domain.
Second, we propose a novel \textbf{M}ultimodal \textbf{F}requency-\textbf{D}riven \textbf{C}hange \textbf{D}etector (\textbf{MFDCD}). Our framework tackles the core challenges by operating in the frequency domain for both visual and textual features. It consists of two specialized components: 
(1) the Dynamic Frequency Coupler (DFC), which utilizes DWT operation to separate low-frequency components (encoding structural continuity) from high-frequency details (capturing transition boundaries). Within the DFC, an Adaptive Sparse Frequency Fusion (ASFF) block  delineates boundaries by dynamically combining high-frequency information with low-level visual features. A Bidirectional Temporal Frequency Fusion (BTFF) block fuses low-level frequency and high-level visual features across time points to model the linear structure of change regions, preventing fragmentation; and 
(2) the Textual Frequency Filter (TFF), which first 
converts textual features into the frequency domain, a graph-based filtering mechanism then acts as a semantic reasoning to extract salient semantic components and joint multimodal representation learning with visual features. The resulting unambiguous features then guide the model toward confident semantic predictions.
Our contributions are as follows:
\begin{itemize}
	\item 
	We introduce RB-SCD, a challenging  benchmark dataset specifically designed for SCD of roads and bridges. It addresses the critical scarcity of domain-specific data by providing 11 fine-grained semantic change categories, enabling a more nuanced and practical analysis of traffic infrastructure evolution.
	\item 
	We propose  MFDCD, a    framework that for the first time integrates multimodal (visual-textual) features in the frequency domain for CD. 
	The framework overcomes conventional limitations through two key components: the DFC leverages wavelet transform to  robustly model linear and heterogeneous semantic changes, and the TFF
	acts as a semantic reasoning engine to resolve high semantic ambiguity by aligning textual concepts with visual patterns in the frequency domain.
	\item 
	Extensive experiments show the state-of-the-art performance of our MFDCD. Specifically, on our proposed RB-SCD dataset, it excels at identifying complex semantic changes in traffic scenes. On three public benchmarks for BCD, it shows superior generalization ability and achieves highly competitive results.
\end{itemize}


\section{Related Work}
\subsection{CD Datasets and Methods}
Existing BCD datasets such as LEVIR~\cite{LEVIR}, WHU~\cite{WHU}, CDD~\cite{CDD}, SYSU~\cite{SYSU} are primarily designed for BCD tasks and cannot provide specific semantic change information. For SCD, datasets like Landsat~\cite{Landsat-SCD}, HRSCD~\cite{HRSCD}, SECOND~\cite{SECOND}  mainly focus on changes in land cover, with little attention paid to the changes in roads and bridges in traffic scenes.
In contrast, our RB-SCD dataset is a high-resolution, fine-grained SCD dataset specifically curated and annotated for road and bridge changes in traffic scenes. 
It enables precise and meaningful evaluation of CD algorithms in traffic scenarios.

In recent years, deep learning has significantly improved the accuracy of CD through its ability to extract relevant features from raw data.
C-3PO~\cite{C-3PO} reframes CD as a semantic segmentation task, proposing a new paradigm that adapts existing powerful segmentation networks for CD by introducing a dedicated module to learn different change types.
S²PNet~\cite{S2PNet} enhances CD in complex VHR scenes by introducing a self-structured feature pyramid for better multi-scale object cognition.
MHF²-Net~\cite{MHFNet} addresses the multi-scale object detection challenge in VHR CD by proposing a hierarchical feature fusion network that explicitly enhances high-frequency details.
AGFormer~\cite{AGFormer} introduces a novel transformer-based model that mitigates classifier bias from class imbalance in CD by using an anchor-guided regularization strategy on a hypersphere to ensure better inter-class separability and intra-class balance.
CdSC~\cite{Csdc} introduces a cross-differential semantic consistency network that explores the  differences in bi-temporal features while maintaining their semantic consistency. 
HGINet~\cite{HGINet} builds a multi-level perception aggregation network with a pyramid architecture to extract features that distinguish different categories across multiple levels.
However, these methods only consider single-modal data, i.e., images, while neglecting the rich semantic information embedded in multimodal data, thus encountering a performance bottleneck.

\subsection{Multimodal Learning for Change Detection}
Recent studies have shown that multimodal information is beneficial for visual image analysis. 
ChangeCLIP~\cite{ChangeCLIP} designs a novel CD framework that utilizes semantic information from image-text pairs and introduce an innovative differential feature compensation module to capture fine-grained semantic changes between them. 
HFA-PANet~\cite{HFA-PANet} addresses the domain shift challenge in multimodal CD by proposing a non-siamese network that hierarchically aligns features from heterogeneous images to effectively capture their differences.
AEKAN~\cite{AEKAN} introduces a novel unsupervised approach for multimodal CD by using a Siamese Kolmogorov-Arnold Network (KAN) AutoEncoder to effectively learn the latent commonality features between heterogeneous images.
CFRL~\cite{CFRL} presents an unsupervised framework for multimodal CD that learns comparable commonality features by cyclically reconstructing, re-encoding, and aligning features from heterogeneous images.
While these pioneering works highlight the growing importance of multimodality, they primarily fall into two categories: visual-textual fusion in the spatial domain or feature alignment for heterogeneous image inputs. Our work carves out a distinct niche by being the first to introduce a visual-textual paradigm that operates within the frequency domain, specifically designed to resolve the deep semantic ambiguities found in complex, heterogeneous RS scenes
\subsection{Frequency-Domain Analysis in Deep Learning}
In image processing, the combination of CNN and frequency domain information has been widely applied in tasks such as image restoration, super-resolution, and semantic segmentation. W-Unet~\cite{Zhao} replaces down-sampling operations with DWT to reduce the impact of image noise while avoiding information loss. FFT-ReLU ~\cite{Mao_Liu_Liu_Li_Shen_Wang_2023} performs image deblurring tasks in the frequency domain based on fast Fourier transform. ARFFT~\cite{Zhu} proposes a spatial-frequency fusion block to enhance the representation capability of Transformers and extend the receptive field across the entire image. XNet~\cite{XNet} utilizes 
high- and low-frequency information to design a framework supporting both fully supervised and semi-supervised tasks. 
\begin{figure*}[!t]
	\centering
	\includegraphics[width=1\textwidth]{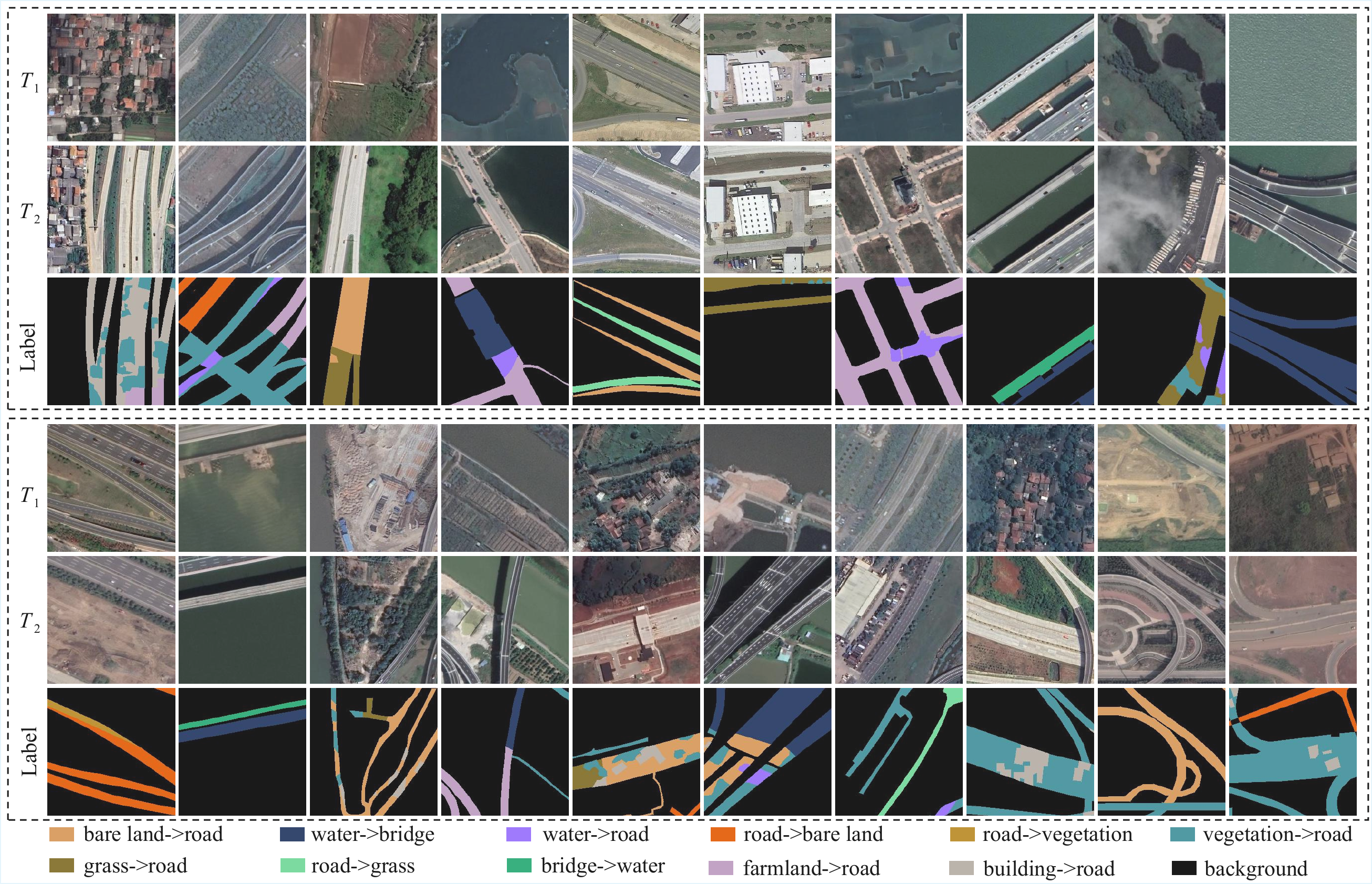} 
	\caption{Annotation examples on  RB-SCD dataset. $T_1$ and $T_2$ represent the pre- and post-change images, respectively. The Label row shows the ground truth.}
	\label{fig:annotate_samples}
\end{figure*}
Given the characteristics of RS images, such as high resolution and susceptibility to spectral variations caused by seasonal changes, using only frequency domain features often results in the loss of critical spatial information. Therefore, in this paper, we  model multimodal features in the frequency domain, effectively balancing the advantages of visual features, textual features, and frequency domain features, achieving promising results.


\section{RB-SCD Dataset}
As established, existing benchmarks are not suited for the  task of road and bridge CD in traffic scenes. Therefore, our primary goal in creating the RB-SCD dataset is to provide detailed, fine-grained annotations that explicitly capture the semantic evolution of linear networks, while also covering a wide range of complex real-world scenes. 
\label{sec 3}
\subsection{Data Annotation and Analysis}

\begin{figure*}[t]
	\centering
	\includegraphics[width=\linewidth]
	{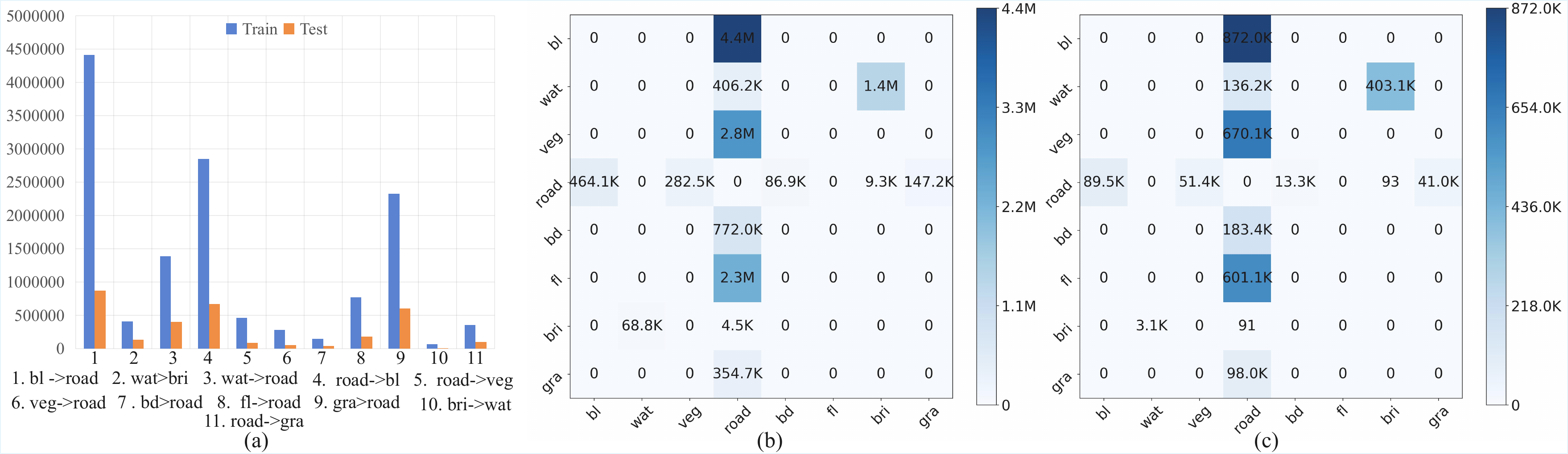} 
	\caption{Statistical analysis of the RB-SCD dataset. (a) Per-class pixel distribution for  semantic change categories.
	(b-c) Semantic transition matrices for the training and test sets, respectively, showing the pixel counts for each from -\textgreater{} to change type. Here, \textit{bl}, \textit{wat}, \textit{veg}, \textit{bd}, \textit{fl}, \textit{bri}, and \textit{gra} denote abbreviations for
	\textit{bare land}, \textit{water}, \textit{vegetation}, \textit{building}, \textit{farmland}, \textit{bridge}, and \textit{grass}, respectively.}
	\label{fig:dataset_Statistics}
\end{figure*}

\textbf{Rich Semantic Annotations.} 
As shown in Fig.~\ref{fig:annotate_samples} and Fig.~\ref{fig:dataset_Statistics}, 
unlike traditional BCD datasets, RB-SCD provides fine-grained annotations for 11 distinct semantic change types. These change annotations are constructed from eight fundamental land-cover classes, which are detailed in Tab.~\ref{tab:subcategories}.
A critical contribution of RB-SCD is its fine-grained semantic granularity, which is not merely a detail but an essential feature for meaningful real-world applications.
This granularity allows for a precise understanding of urban evolution. For instance, categories like \textit{farmland-\textgreater{}road} and \textit{vegetation-\textgreater{}road} directly track the impact of urbanization on
\begin{table}[]
	\centering
	\caption{Specific subcategories of land-cover classes in RB-SCD dataset.}
	\resizebox{1\textwidth}{!}{
		\begin{tabular}{ccccc}
			\toprule
			Categories & \multicolumn{4}{c}{Description}    
			\\ \midrule
			Road             & \multicolumn{4}{l}{street, highway, dirt road, walkway, railway,  etc.}                                                                        \\
			\cline{2-5}
			Bridge           & \multicolumn{4}{l}{\begin{tabular}[c]{@{}l@{}}beam bridge, arch bridge,  cable-stayed bridge, suspension bridge, etc.\end{tabular}}     \\
			\cline{2-5}
			Bare land        & \multicolumn{4}{l}{\begin{tabular}[c]{@{}l@{}}wasteland, sand patches area, unpaved open grounds, road or railway construction zones, etc.\end{tabular}}                       \\
			\cline{2-5}
			Grass           & \multicolumn{4}{l}{\begin{tabular}[c]{@{}l@{}}urban lawn, green belt, wetland grass, riparian  grass zone, meadow etc.\end{tabular}}                                                                                                      \\
			\cline{2-5}
			Water            & \multicolumn{4}{l}{\begin{tabular}[c]{@{}l@{}}ocean, lake, river, reservoir, pond, etc.\end{tabular}}                                                                           \\
			\cline{2-5}
			Farmland         & \multicolumn{4}{l}{agricultural land (with and without crops).}                                                                 \\
			\cline{2-5}
			Vegetation       & \multicolumn{4}{l}{forest, shrub, urban vegetation buffer,  etc.}                                                                                            \\
			\cline{2-5}
			Building         & \multicolumn{4}{l}{\begin{tabular}[c]{@{}l@{}}residential building, commercial building,  industrial building, under-construction building\\ (with clear structure), etc. 
			\end{tabular}}
			\\ \bottomrule
	\end{tabular}}
	\label{tab:subcategories}
\end{table}
\begin{figure*}[t]
	\centering
	\includegraphics[width=1\linewidth]{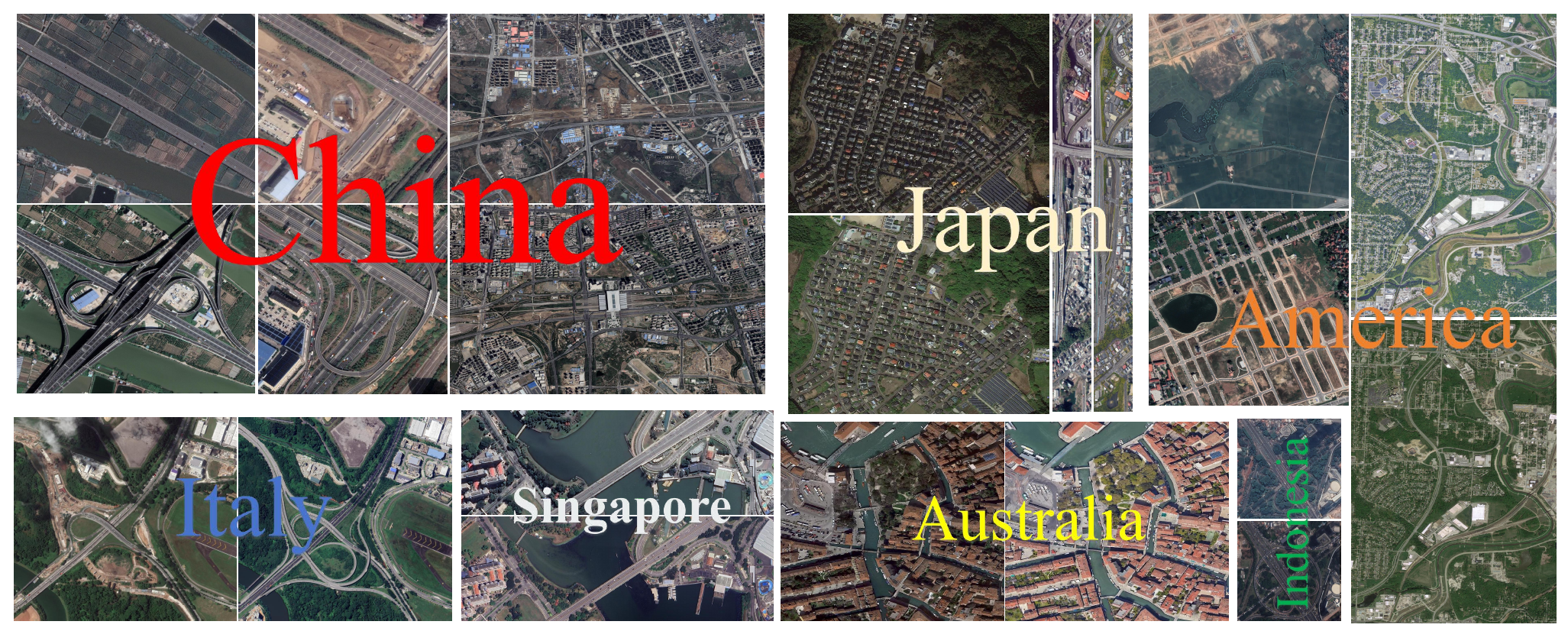} 
	\caption{
		The diverse geographic distribution of the RB-SCD dataset. The samples span multiple countries and continents, including varied urban and rural landscapes from China, Japan, America, Italy, Singapore, Australia, and Indonesia. This diversity ensures the dataset's robustness and challenges models to generalize across different environmental conditions and infrastructure styles. }
	\label{fig:data_samples}
\end{figure*}
\begin{figure}[t!]
	\centering
	\includegraphics[width=1\linewidth]{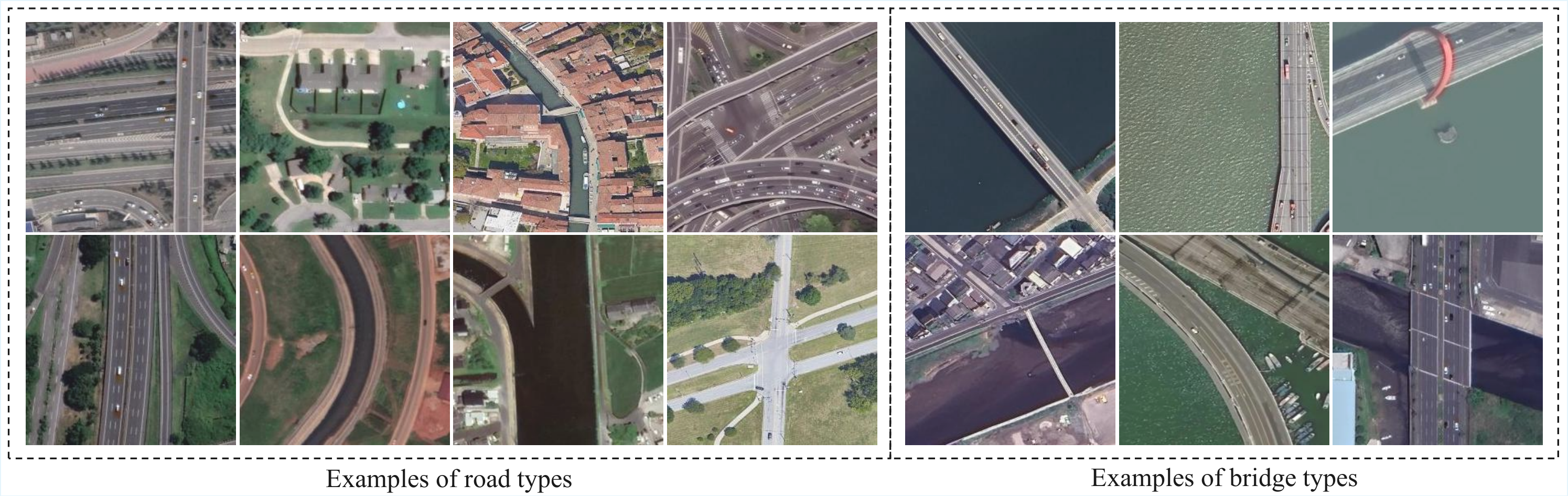} 
	\caption{Examples of road and bridge types, where the first four columns are roads, and the last three columns are bridges. }
	\label{fig:roads_and_bridges}
\end{figure}
ecological and agricultural land, providing critical data for environmental assessment. Conversely, changes such as \textit{road-\textgreater{}grass} or the rare but critical \textit{bridge-\textgreater{}water} capture infrastructure degradation or demolition, which is vital for safety monitoring and disaster response. Distinguishing between road widening (an implicit category captured through pixel-level changes) and \textit{building-\textgreater{}road} is crucial for urban renewal analysis and traffic management. 
All annotations are performed by a professional team in RS and underwent rigorous quality control to ensure accuracy and semantic consistency.
It is worth noting that some changes, such as bridges directly turning into roads, are rarely observed in reality. Additionally, other types of changes, such as those between bridges and some other categories, occur in only a few samples, and we set their pixel number to 0 in the training process.

\textbf{Diverse Geographic Coverage.} The image pairs are sourced from Google Earth, with a temporal coverage spanning from 2007 to 2022. As shown in Fig.~\ref{fig:data_samples}, following the principles in~\cite{Benchmark_Dataset}, we collect 260 pairs of RS images with resolutions ranging from 270 × 534 to 7,215 × 4,366 (0.59 m/pixel), which can be cropped into 5,016 pairs of 256 × 256 patches, including 4,015 pairs for the training set and 1,001 pairs for the testing set.
The dataset covers various regions worldwide, including Chinese cities such as Hefei, Xi’an, Chongqing and Guangzhou, as well as representative areas from Japan, Singapore, Nigeria, Italy,
America, and Indonesia, etc. This extensive geographic coverage ensures that the dataset encompasses traffic scenes under varying climatic conditions, geological structures, and urban planning contexts, thereby guaranteeing its diversity and representativeness.

\textbf{Multiple Road and Bridge Types.}
As shown in Fig.~\ref{fig:roads_and_bridges}, the dataset covers diverse road and bridge types across varied regions and urban contexts. Road categories range from urban main roads, secondary streets, and highways to elevated roads designed for congestion relief, while bridge types include large cable-stayed bridges, girder bridges, and pedestrian bridges. This diversity captures different engineering styles and traffic environments, providing robust support for training CD models across multiple scales and diverse scenarios.

\subsection{Dataset Limitations}
As shown in Fig.~\ref{fig:dataset_Statistics}, RB-SCD inevitably exhibits class imbalance across semantic change types. Common transitions (e.g., road widening, bridge construction) are well represented, while rarer cases (e.g., road repair, demolition) appear less frequently. We have used a weighted cross-entropy loss to mitigate the class imbalance issue in the RB-SCD dataset, which assigns higher weights to the losses of under-represented classes.
We will further collect more samples of different types of changes in future work and employ techniques such as data augmentation and sample weighting to alleviate the model's difficulty in detecting rare changes.
\begin{figure*}[t]
	\centering
	\includegraphics[width=\linewidth]{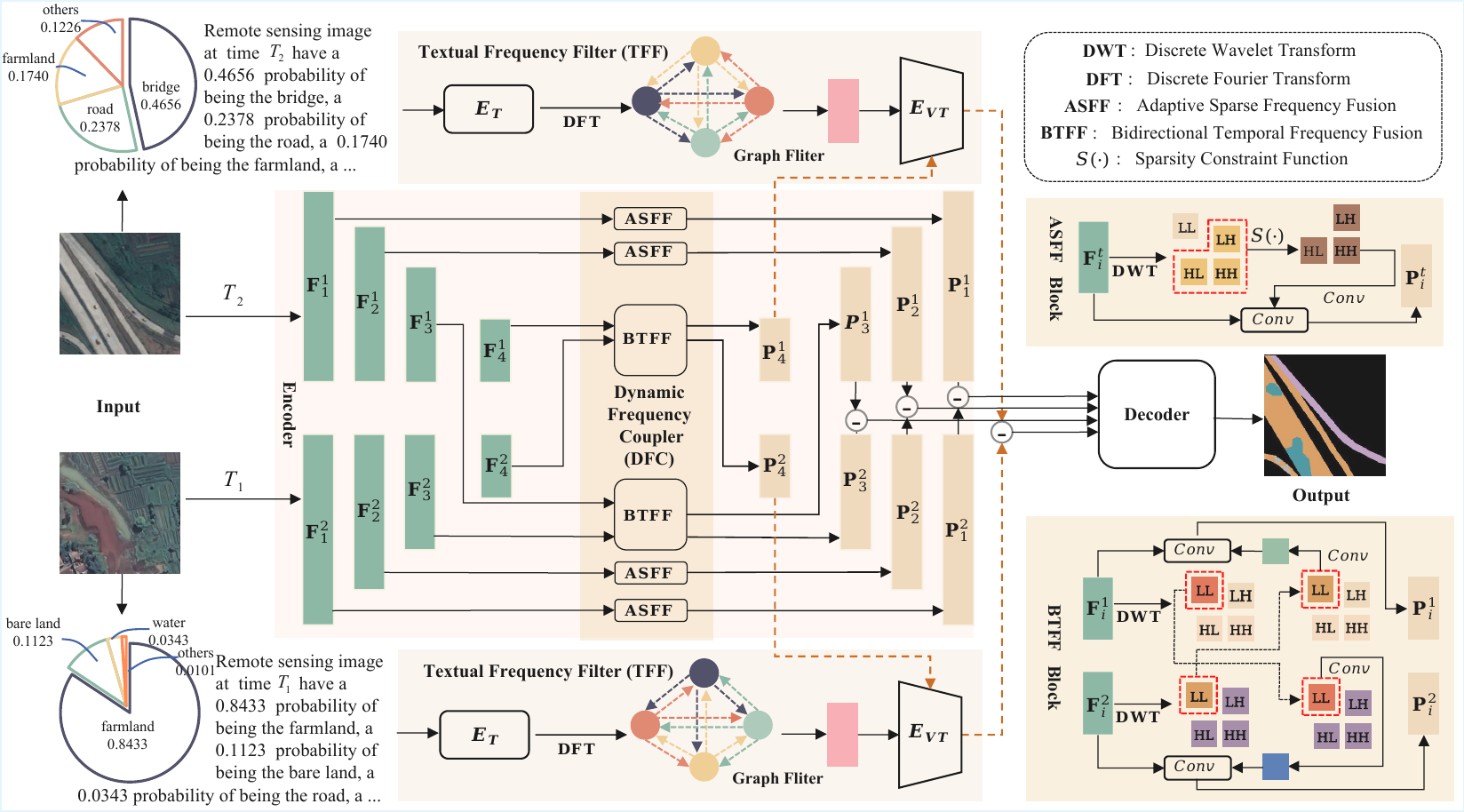}
	\caption {
		Overview of our proposed Multimodal Frequency-Driven Change Detector (MFDCD). The framework takes bi-temporal images ($T_1,T_2$) as input. The Dynamic Frequency Coupler (DFC) processes visual features from a ResNet50~\cite{ResNet50} backbone to robustly model linear and heterogeneous transitions. It employs an ASFF block to precisely delineate change boundaries and a BTFF block to model the overall change evolution, preventing fragmentation. Concurrently, the Textual Frequency Filter (TFF) processes CLIP-inferred textual descriptions to resolve high semantic ambiguity. It transforms text into the frequency domain and uses a graph filter to reason about the most relevant semantic concepts. Finally, the enhanced visual features from the DFC, guided by the semantic features from the TFF, are fed into a decoder to generate the final semantic change map.	
	}
	\label{fig:overview}
\end{figure*}
\section{Methodology }
\label{method}
\subsection{Overview }
As shown in Fig. \ref{fig:overview}, 
our MFDCD follows an encoder-decoder architecture. Given the bi-temporal images \( T_1 \) and \( T_2 \), we first employ a ResNet50~\cite{ResNet50} backbone to extract multi-scale visual features  \( \mathbf{F}_i^t \in \mathbb{R}^{B \times C_i \times H_i \times W_i} (i \in \left \{ 1,2,3,4 \right \}, t\in \left \{ 1,2 \right \})\). The core innovation consists of two parallel branches: a Dynamic Frequency Coupler (DFC) for  linear and heterogeneous semantic change modeling and a Textual Frequency Filter (TFF) for semantic disambiguation. Finally, the enhanced features from these two branches are fused and passed to the decoder to generate the final semantic change map.
\subsection{Dynamic Frequency Coupler (DFC)}
To address the challenge of linear and heterogeneous semantic change transitions, we first employ the DWT operation. Unlike standard convolutions, as shown in Fig.~\ref{fig:Frequency_analysis}(a), DWT naturally decomposes features into two components that are crucial for this task: 
high-frequency components, which are sensitive to the detailed boundaries where heterogeneous land-cover transitions occur, and low-frequency components, which are ideal for capturing the long-range continuity and global structure of linear targets like roads.
However, merely separating these frequency bands is insufficient; they must be intelligently integrated back into the feature hierarchy. Therefore, as shown in Fig.~\ref{fig:overview},  DFC incorporates two specialized fusion sub-modules:
the Adaptive Sparse Frequency Fusion (ASFF) and the Bidirectional Temporal Frequency Fusion (BTFF) block.

\textbf{Adaptive Sparse Frequency Fusion (ASFF).} ASFF block is designed for precise boundary delineation. It achieves this by fusing the distilled high-frequency details with the rich, fine-grained low-level spatial features from the backbone. 
Low-level features are rich in detailed information, while high-frequency features contain information about boundaries and abrupt changes, which naturally correspond to low-level features. 
Let $\mathbf{F}_i^t \in \mathbb{R}^{B\times C_i \times H_i \times W_i} (i \in \left \{1, 2\right \} , t \in \left \{1, 2\right \})$ be the low-level visual features from layer $i$ at time $t$. We first apply the DWT  to obtain the  frequency sub-bands:
\begin{align}
\mathbf{LL}_i^t, \mathbf{HL}_i^t, \mathbf{LH}_i^t, \mathbf{HH}_i^t=\mathcal{W}(\mathbf{F}_i^t),
\end{align} 
where $\mathbf{LL}_i^t$ is the low-frequency band, $\mathbf{HL}_i^t, \mathbf{LH}_i^t,$ and $\mathbf{HH}_i^t$ signifying the high-frequency bands. $\mathcal{W}(\cdot)$ is the DWT operation. 
The high-frequency sub-bands are then fused into a unified high-frequency representation:
\begin{equation}
\mathbf{H}_i^t = C_{fuse}(Cat(\mathbf{HL}_i^t, \mathbf{LH}_i^t, \mathbf{HH}_i^t)),
\end{equation}
where $C_{fuse}(\cdot)$ is a convolutional block and $Cat(\cdot)$ is the concatenation operation.
However, high-frequency features are highly susceptible to noise. Especially in the early stages of training, when the model is not yet well-optimized, these high-frequency features often contain significant clutter and irrelevant details.
Therefore, we design a principled, sparsity-aware gating mechanism.
$S(\cdot)$ is defined as a dynamic sparsity constraint function,  which adaptively thresholds the high-frequency features based on the training progress:
\begin{equation}
\tilde{\mathbf{H}}_i^t = S(\mathbf{H}_i^t, \lambda (p), g) = 
\begin{dcases} 
\medspace\medspace \mathbf{H}_i^t, & {if } \quad |\mathbf{H}_i^t| > g \cdot \lambda(p) \\
\medspace\medspace\medspace  \mathbf{0}, & \quad\quad{otherwise}
\end{dcases},
\end{equation}
where $\lambda(p)= \lambda_0 e^{-\delta    p}$ is the dynamic sparsity strength, $\lambda_0$ is the initial sparsity strength, 
$\delta$ is the sparsity decay rate,  $p$ is the current training iteration, and $g$ is a constant controlling the sparsity strength.
Early in training, a strong constraint (large $\lambda(p)$) allows only the most salient boundary features to pass, ensuring stable learning. As training progresses, the constraint $\lambda(p)$ decays, gradually permitting finer details to be fused for precise boundary refinement. This coarse-to-fine approach is more robust than a fixed or learnable threshold, adapting to the model's learning stage.
Finally, we treat this refined high-frequency map as a spatially-aware gate to modulate the low-level features. The final fused feature representation $\mathbf{P}_i^t$ is generated by a fusion block $F_{\text{ASFF}}$:
\begin{equation}
\mathbf{P}_i^t = F_{\text{ASFF}}(\mathbf{F}_i^t, \tilde{\mathbf{H}}_i^t) = C_{out}(Cat(Up(\tilde{\mathbf{H}}_i^t), \mathbf{F}_i^t)),
\end{equation}
where $Up(\cdot)$ is a bilinear upsampling operator to match spatial dimensions, and $C_{out}$ is the output convolution layer.

\textbf{Bidirectional Temporal Frequency Fusion (BTFF).}
To robustly model the change evolution of linear structures and prevent fragmentation, the model must explicitly compare the structural information across two time points. The BTFF module is designed to achieve this via a cross-temporal structure injection mechanism. The core principle is to directly infuse the low-frequency structural representation from one time point into the high-level feature space of the other, thereby forcing the network to reason about semantic features in the context of the opposing temporal structure.

Let $\mathbf{F}_i^t \in \mathbb{R}^{B\times C_i \times H_i \times W_i} (i \in \left \{3, 4\right \} , t \in \left \{1, 2\right \})$  be the high-level features and $\mathbf{LL}_i^t$ be the low-frequency structural components at time $t$ for layer $i$. To compute the temporally-informed feature representation $\mathbf{P}_i^1$ for time $t=1$, we inject the structural information from time $t=2$.
The low-frequency component from the second time point, $\mathbf{LL}_i^2$, represents a structural prior. To prepare this prior for fusion, it must be projected into a higher-dimensional space compatible with the semantic features. This is achieved by a projection function $\Phi_L(\cdot)$, composed of a convolutional block followed by an upsampling operator $Up(\cdot)$:
\begin{equation}
\widehat{\mathbf{LL}}_i^2 = \Phi_L(\mathbf{LL}_i^2) = Up(C_{proj}(\mathbf{LL}_i^2)),
\end{equation}
where $C_{proj}(\cdot)$ maps the low-frequency features to an intermediate representation.

Next, the semantic feature space of the first time point is augmented by the structural prior from the second. We define an augmentation operator $\Psi (\cdot, \cdot)$ which concatenates the two representations and applies a non-linear fusion transformation $C_{trans}(\cdot)$ 
to model their complex interactions: 
\begin{equation}
\mathbf{P}_i^1 = \Psi (\mathbf{F}_i^1, \widehat{\mathbf{LL}}_i^2) = C_{trans}(Cat(\mathbf{F}_i^1,\widehat{\mathbf{LL}}_i^2))+ \mathbf{F}_i^1,
\end{equation}
where   $\mathbf{P}_i^1$ represents the learned structural difference and contextual information derived from the cross-temporal comparison.
The process is performed symmetrically to compute $\mathbf{P}_i^2$ by injecting $\mathbf{LL}_i^1$ into $\mathbf{F}_i^2$, thus creating a bidirectional flow of structural information:
\begin{equation}
\mathbf{P}_i^2 = \Psi (\mathbf{F}_i^2, \widehat{\mathbf{LL}}_i^1) = C_{trans}(Cat(\mathbf{F}_i^2,\widehat{\mathbf{LL}}_i^1))+ \mathbf{F}_i^2.
\end{equation}

This mechanism ensures that the model's understanding of each time point is conditioned on the structural reality of the other, making it highly effective at modeling  change regions.

\subsection{Textual Frequency Filter (TFF)}
\begin{figure*}[t]
	\centering
	\includegraphics[width=1\linewidth]{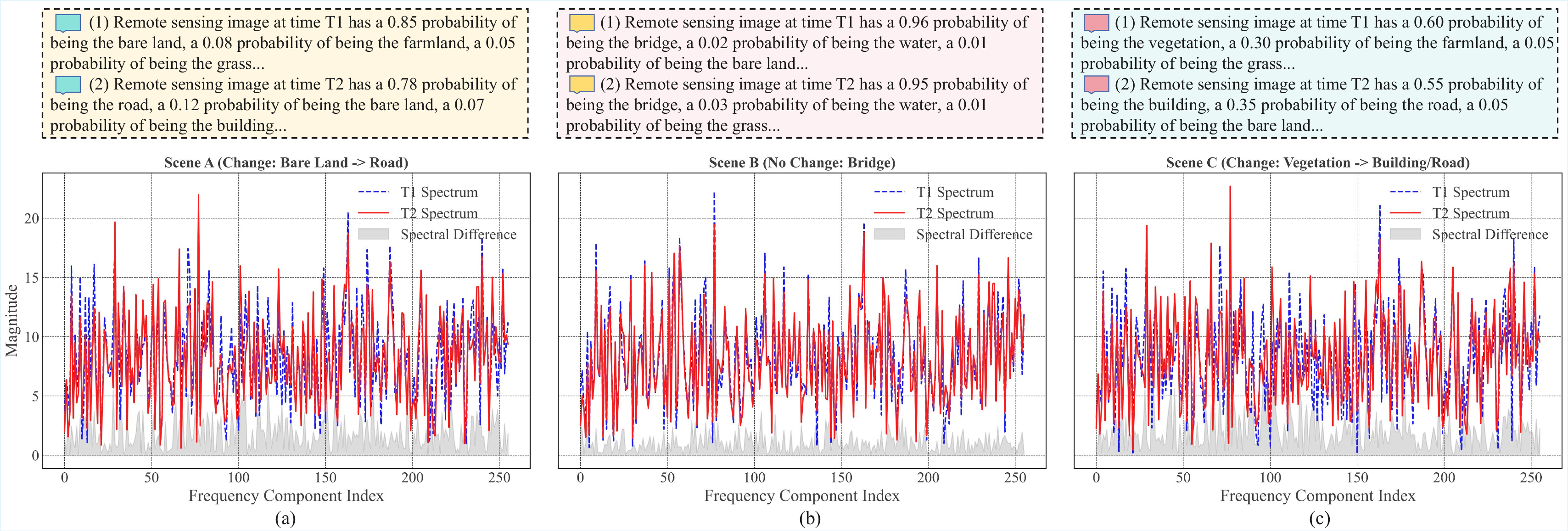} 
	\caption{Visualization of text embedding spectra in the frequency domain by using the  discrete Fourier transform.  
		 The blue dashed line represents the spectrum of the $T_1$ image description, while the red solid line is the $T_2$ description. The gray shaded area highlights the absolute difference between the two spectra. 	
}
	\label{fig:TFF_Frequency_Analysis}
	
\end{figure*}

While recent vision-language models like ChangeCLIP~\cite{ChangeCLIP} leverage textual priors, their fusion in the noisy spatial-domain still struggles to resolve deep semantic ambiguities. To overcome this, our Textual Frequency Filter (TFF) operates directly in the frequency domain to achieve a more robust semantic alignment. The core idea is to transform textual embeddings into a global signature that separates stable semantic concepts from high-frequency noise, thus providing clearer guidance to disambiguate challenging visual scenes. 
Our empirical validation in Fig.~\ref{fig:TFF_Frequency_Analysis} confirms this insight. We visualize the magnitude spectra of CLIP-generated text embeddings for representative change and no-change scenarios. In the no-change scene (Fig.~\ref{fig:TFF_Frequency_Analysis}(b)), the $T_1$ and $T_2$ spectra are nearly identical, demonstrating the stability of the spectral representation. Conversely, in the change scenes (Fig.~\ref{fig:TFF_Frequency_Analysis}(a) and ~\ref{fig:TFF_Frequency_Analysis}(c)), the spectra diverge significantly, as highlighted by the large spectral difference (gray area). This demonstrates that semantic change is effectively encoded as a quantifiable signal in the frequency domain.
Therefore, we employ the DFT as the foundational tool within our TFF. 

As shown in Fig.~\ref{fig:overview}, we first leverage the powerful zero-shot inference capability of the CLIP~\cite{CLIP} model to obtain textual information from RS images for multimodal modeling. For the categories inferred by the CLIP model, we select the top five categories with the highest probability values. For the foreground features, the text description \( T_d \) is designed in the format  ``Remote sensing image at  time $ \left \{ period  \right \}$  has a $ \left \{ value  \right \}$  probability of being the $ \left \{ key  \right \}$ '' where $ \left \{ period  \right \}$ represents either $T_1$ or $T_2$, $ \left \{ value  \right \}$ is the probability value of the corresponding category and $ \left \{ key  \right \}$ represents the category itself. We  leverage a pre-trained CLIP text encoder~\cite{CLIP}, denoted as $E_T$, to obtain token-level embeddings for each textual description $T_d$. 
\begin{equation}
\mathbf{E}^t=E_T(T_d).
\end{equation}

This process yields a feature tensor $\mathbf{E}^t \in \mathbb{R}^{B \times N \times D}$, where $B$ is the batch size, $N$ is the sequence length (number of tokens), and $D$ is the feature dimension for each token.

To transform this sequence of token features into the frequency domain, we apply the DFT along the sequence dimension $N$. This operator, denoted as $\mathcal{F}$, analyzes the semantic patterns across the entire sequence. For each sample $b$ in the batch and each feature dimension $d$, the $k$-th component of the resulting frequency spectrum $\mathbf{T}_{{freq}} \in \mathbb{C}^{B \times N \times D}$ is computed as:
\begin{equation}
\mathbf{T}_{{freq}}(b, k, d) = \mathcal{F}(\mathbf{E}^t)(b, k, d) = \sum_{n=0}^{N-1} \mathbf{E}^t(b, n, d) \cdot e^{-j2\pi kn/N},
\end{equation}
where $n$ is the index over the sequence of tokens, $n \in \{0, \dots, N-1\}$, $k$ is the index for the frequency components, $k \in \{0, \dots, N-1\}$, $d$ is the index for the feature dimensions, $d \in \{0, \dots, D-1\}$, $j$ is the imaginary unit.
The resulting $\mathbf{T}_{{freq}}$ is a complex-valued tensor that encodes how semantic features are distributed across different frequencies of the text sequence. This allows us to capture global relational patterns and periodic structures, yielding a more robust representation for subsequent reasoning.

However, these frequency components are not independent. Complex semantic concepts are often defined by global and potentially non-local relationships across the spectrum. To capture these dependencies, we model the frequency components as nodes and introduce a graph filter bank to act as a semantic reasoning engine. 
Specifically, we employ a fully connected graph, a design choice that imposes the weakest structural prior. 
This is crucial because key semantic concepts (e.g., \textit{bare land} and \textit{road}) often have strong, non-local relationships within our structured text descriptions, which would be overlooked by models assuming locality, such as sparse graphs based on token adjacency.
This maximally flexible structure allows our graph filter bank to learn arbitrary combinations of frequencies that characterize specific change types, without being prematurely constrained by locality assumptions. 
For the frequency feature \( \mathbf{T}_{freq} \), we split it into real and imaginary parts and concatenate them to construct node features. The feature matrix \( \mathbf {X}  \) is the collection of features for all nodes, which allows us to convert frequency domain information into node features for the graph:
\begin{equation}
\mathbf {X}=Cat(e(\mathbf{T}_{freq}   ), Im(\mathbf{T}_{freq})  ), \mathbf {X} \in \mathbb{R}^{B\times N\times 2C},
\end{equation}
where $Re(\mathbf{T}_{freq}   ) $ and $Im(\mathbf{T}_{freq})$ represent the real and imaginary parts of the frequency domain features, respectively. The edge set $\mathbf {E} $  can be constructed by combining each pair of nodes:
\begin{equation}
\mathbf {E}=\left \{ (i,j)\mid 1\le i\ne j \le N \right \},  \mathbf {E} \in\mathbb{Z}^{2\times \frac{N(N-1)}{2} }.  
\end{equation} 

The graph filter bank is  denoted as $\mathcal{K} = \{k_1, k_2, \dots, k_L\}$ and implemented as a set of $L$ parallel  layers, $\{\mathcal{G}_l\}_{l=1}^L$, where each layer $\mathcal{G}_l(\cdot)$ corresponds to a filter $k_l$. 
Unlike a 1D CNN that captures only local dependencies, a graph filter performs relational reasoning over the entire spectral structure. Its message-passing mechanism is designed to learn a filter function that enhances or suppresses global spectral patterns associated with specific semantic changes. 
Each filter then is parameterized by its own weight matrix $\mathbf {W}_l$ and learns to capture a specific type of semantic relationship from the node features $\mathbf {X}$, thereby obtaining salient features related to the changes:
\begin{equation}
\mathbf {X}'_l= \mathcal{G}_l(\mathbf {X}, \hat{\mathbf {A}   } ) =\sigma(\hat{\mathbf {A}   }\mathbf {X}\mathbf {W}_l  ),
\end{equation}
where $\mathbf {X}'_l$ represents the node features after the 
$l$-th filter, $\hat{\mathbf {A}}$ is the normalized adjacency matrix derived from $\mathbf {E}$ and  
\( \sigma \) is the activation function.
The outputs from all filters in the bank $\mathcal{K}$ are then aggregated to form a comprehensive, refined semantic representation $\mathbf{X}_T$:
\begin{equation}
\mathbf{X}_T = \sum_{l=1}^{L} \mathbf {X}'_l = \sum_{l=1}^{L} \mathcal{G}_l(\mathbf {X}'_l, \hat{\mathbf{A}}).
\end{equation}

Let $V_{ct}$ be the spatial visual context tokens from the image backbone. A context decoder ${E}_{\text{VT}}$, aligns the refined frequency-domain semantics $\mathbf{X}_T$ with this spatial visual evidence to compute a visually-informed  feature:
\begin{equation}
\mathbf{X}_T^{refined} = {E}_{\text{VT}}(V_{ct}, \mathbf{X}_T).
\label{eq:context_decoder}
\end{equation}

By analyzing textual semantics in the frequency domain and aligning them with spatial visual features, TFF effectively resolves semantic ambiguities and provides reliable guidance for change prediction.

\section{Experiments and Results}

\subsection{CD Datasets}
In addition to using our proposed dataset,   we have employed three public CD datasets to evaluate the robustness of our model. 
The LEVIR-CD~\cite{LEVIR}  dataset consists of 637 pairs of 1,024 × 1,024 satellite images with a spatial resolution of 0.5 m/pixel. It primarily focuses on building area changes,  and is split into 7,120/1,024/2,048 pairs for training, validation, and testing.
The SYSU-CD~\cite{SYSU} dataset is a collection of 20,000 pairs of 256 × 256 high-resolution aerial images, also at 0.5 m/pixel resolution. Captured in the Hong Kong area, it includes diverse changes like urban construction and suburban expansion, and is divided into 12,000/4,000/4,000 pairs for train/val/test.
The WHU-CD~\cite{WHU} dataset comprises a pair of large-format aerial images with a size of 32,507 × 15,354 and a higher resolution of 0.2 m/pixel. It covers areas before and after an earthquake, primarily including building renovations and additions. Following the standard split, it provides 5,947/743/744 image patches.

\subsection{Implementation Details}
Our experiments are conducted on a single  GeForce RTX 3090 GPU. The AdamW is employed, with a learning rate of 0.0001 and a weight decay parameter of 0.01. 
 The maximum number of iterations is set to 40,000. For data augmentation, RandomFlip and PhotoMetricDistortion are applied. 
The constant \( g\) in the ASFF block is set to 0.1, the decay rate $\delta $ is set to 0.0001, and \( \lambda_0 \) is set to 1.
We use a weighted cross-entropy loss to mitigate the class imbalance issue in the RB-SCD dataset, which assigns a weight of 1.5 to the losses of under-represented classes, while more common classes retain a weight of 1.0, thereby increasing the penalty for misclassifying rarer change types.
To evaluate the performance of our experiments, we adopt four widely used metrics: recall (Rec), precision (Pre), F1-score (F1) and intersection over union (IoU), where IoU and
F1 are the primary metrics, while the others are auxiliary metrics. 

\begin{table*}[t]
	\caption{Comparison of quantitative performance for semantic change detection on RB-SCD dataset. All metrics are expressed as percentages (\%). Results highlighted in \textcolor{red}{red} indicate the best performance, while those in \textcolor{blue}{blue} denote the second-best performance. }
	\centering
	\setlength{\tabcolsep}{1mm}
	\resizebox{1\textwidth}{!}{
		\begin{tabular}{c|c|c|c|c|c}
			\toprule
			\multirow{2}{*}{Class}   & HGINet~\cite{HGINet}                            & CdSC~\cite{Csdc}                           & DEFO~\cite{DecoderFusion}      &ChangeCLIP~\cite{ChangeCLIP} &Ours                        \\
			& IoU\quad F1\quad Rec\quad Pre                & IoU\quad F1\quad Rec\quad Pre              & IoU\quad F1\quad Rec\quad Pre& IoU\quad F1\quad Rec\quad Pre  & IoU\quad F1\quad Rec\quad Pre         \\
			\midrule
			background  
			&97.76\quad98.87\quad99.14\quad98.60      
			&{\color{blue}98.22\quad99.10}\quad{\color{red}99.45}\quad98.75         
			&98.18\quad99.08\quad98.27\quad{\color{blue}98.90}   
			&98.11\quad99.05\quad{\color{blue}99.37}\quad98.73         & {\color{red}98.28\quad  99.13}\quad99.34\quad{\color{red}98.93}         \\
			bare land -\textgreater{}road          
			& 45.24\quad62.30\quad61.10\quad63.54          & 48.42\quad65.25\quad63.40\quad67.21    & 50.93\quad67.49\quad{\color{blue}66.04}\quad{\color{red}69.00}        &{\color{blue}51.75\quad68.21}\quad64.87\quad64.87         & {\color{red}52.76\quad69.08}\quad{\color{red}69.49}\quad{\color{blue}68.67}          \\
			water -\textgreater{}bridge            &80.20\quad89.01\quad85.60\quad{\color{blue}92.71}           
			& 85.33\quad92.09\quad{\color{blue}93.32}\quad90.88      
			& 83.10\quad 90.77\quad91.38\quad90.17         
			& {\color{blue}85.77\quad92.34}\quad88.22\quad{\color{red}96.87} 
			&{\color{red}86.83\quad92.95}\quad{\color{red}94.28}\quad91.67         \\
			water-\textgreater{}road              
			& 58.09\quad73.49\quad{\color{blue}76.88}\quad70.39                  
			& 68.64\quad81.49\quad74.82\quad{\color{red}89.26}      
			& 62.78\quad77.14\quad76.21\quad78.88       
			& {\color{blue}68.11\quad81.03}\quad76.23\quad86.47        
			&{\color{red}71.68\quad83.51\quad79.63}\quad{\color{blue}87.78}          \\
			road-\textgreater bare land            
			&33.83\quad50.55\quad40.13\quad68.29   
			&38.33\quad55.42\quad44.58\quad73.22      
			&38.14\quad55.22\quad{\color{red}48.02}\quad64.94          
			&{\color{red}40.57\quad57.72}\quad{\color{blue}47.06}\quad{\color{blue}74.63}      
			&{\color{blue}39.48\quad56.61}\quad43.88\quad{\color{red}79.73}          \\
			road-\textgreater{}vegetation          
			&10.69\quad19.32\quad12.56\quad41.82                 
			&13.65\quad24.03\quad15.25\quad{\color{blue}56.57}    
			&{\color{blue}17.02\quad29.09}\quad{\color{red}22.47}\quad41.25          
			&16.37\quad28.13\quad19.35\quad51.49        
			&{\color{red}20.31\quad33.76}\quad{\color{blue}21.83}\quad{\color{red}74.39}\\
			vegetation-\textgreater{}road         
			&52.61\quad68.94\quad63.10\quad75.98                 
			&{\color{blue} 57.93\quad73.36}\quad69.89\quad{\color{blue}77.19}      
			& 57.84\quad73.29\quad72.30\quad74.41          
			& 57.10\quad72.69\quad68.55\quad{\color{red}77.37}         
			&{\color{red}59.35\quad74.49\quad72.38}\quad76.72       \\
			building-\textgreater{}road            
			&36.56\quad55.57\quad43.44\quad69.78   
			&36.10\quad53.05\quad41.28\quad74.18      
			&39.83\quad56.97\quad{\color{blue}51.63}\quad63.54         
			&{\color{blue}40.46\quad57.61}\quad44.69\quad{\color{red}81.04}         
			& {\color{red}45.44\quad62.49\quad53.06}\quad{\color{blue}75.98}               \\
			farmland-\textgreater{}road               
			& 63.36\quad73.55\quad77.80\quad77.34                 
			&63.80\quad77.90\quad70.78\quad{\color{red}86.61}      
			& {\color{blue} 65.38\quad79.07}\quad75.87\quad82.55          
			& 65.02\quad78.80\quad{\color{red}81.30}\quad76.45          
			&{\color{red}68.40\quad81.24}{\color{blue}\quad79.21\quad83.38}  \\
			grass-\textgreater{}road                  
			&36.55\quad53.54\quad39.09\quad84.94                  
			&36.66\quad53.65\quad46.45\quad64.48      
			&{\color{blue}40.60\quad57.75}\quad{\color{red}48.26}\quad71.89         
			&36.55\quad53.54\quad39.09\quad{\color{red}84.94}        
			&{\color{red}42.49\quad59.64}\quad{\color{blue}47.24\quad80.86}        \\
			road-\textgreater{}grass                   & 22.44\quad36.65\quad{\color{red}32.38}\quad42.21                  
			&15.53\quad26.89\quad17.41\quad58.97     
			&18.41\quad31.10\quad22.75\quad49.14           
			&{\color{red}21.76\quad35.75}\quad{\color{blue}24.21}\quad{\color{blue}68.26}       & {\color{blue}21.65\quad35.59}\quad22.86\quad{\color{red}80.30}   \\
			bridge-\textgreater{}water   
			&46.51\quad63.49\quad49.92\quad87.19                 
			&70.36\quad82.60\quad74.25\quad{\color{red}93.06}       
			&{\color{blue}83.40\quad90.95}\quad{\color{red}93.45}\quad88.57         
			&77.86\quad87.55\quad{\color{blue}90.44}\quad84.85         &{\color{red}83.58\quad91.05}\quad90.34\quad{\color{blue}91.78}\\
			\midrule
			mIoU                          & 48.65                      & 52.75 & 54.63       &{\color{blue}54.95  }    & {\color{red}57.52}\\
			\bottomrule
	\end{tabular}}
	
	\label{tab:RB-SCDD}
\end{table*}
\subsection{Quantitative Assessment Results}
\textbf{Results for Semantic Change Detection on RB-SCD Dataset.} 
Tab.~\ref{tab:RB-SCDD} presents the comparative results of our method and several advanced SCD methods across multiple evaluation metrics.
Across this diverse range of semantic changes, our method achieves significant advantages in both the IoU and F1 metrics. Specifically, for the easily detectable categories like \textit{water-\textgreater{}bridge}, \textit{bridge-\textgreater{}water}, \textit{water-\textgreater{} road} and \textit{farmland-\textgreater{}road}, our model attains IoU scores of 86.83\%, 83.58\%, 71.68\%, and
\begin{table}[t]
	\caption{Performance of the RS foundation model LWGANet on RB-SCD dataset.
	}
	\centering
	\scalebox{0.75}{
		\begin{tabular}{c|cccc|cccc}
			\toprule
			\multirow{2}{*}{Class}        & \multicolumn{4}{c|}{LWGANet-L1~\cite{LWGANet}}                                                                         & \multicolumn{4}{c}{LWGANet-L2~\cite{LWGANet}}                                                                         \\
			& \multicolumn{1}{c}{IoU} & \multicolumn{1}{c}{F1} & \multicolumn{1}{c}{Pre} & \multicolumn{1}{c|}{Rec} & \multicolumn{1}{c}{IoU} & \multicolumn{1}{c}{F1} & \multicolumn{1}{c}{Pre} & \multicolumn{1}{c}{Rec} \\
			\midrule
			background                    & 98.30                    & 99.14                  & 98.84                    & 99.44                    & 98.27                   & 99.13                  & 98.8                     & 99.45                    \\
			bare land -\textgreater{}road              & 54.17                   & 70.27                  & 73.28                    & 67.50                     & 53.17                   & 69.42                  & 74.46                    & 65.02                    \\
			water -\textgreater{}bridge                & 87.29                   & 93.21                  & 92.52                    & 93.92                    & 85.85                   & 92.39                  & 90.95                    & 93.87                    \\
			water-\textgreater{}road                   & 69.53                   & 82.03                  & 82.51                    & 81.55                    & 63.42                   & 77.62                  & 83.84                    & 72.25                    \\
			road-\textgreater{} bare land              & 38.54                   & 55.64                  & 78.94                    & 42.95                    & 38.06                   & 55.14                  & 74.67                    & 43.71                    \\
			road-\textgreater{}vegetation              & 15.54                   & 26.89                  & 55.44                    & 17.75                    & 13.70                    & 24.10                   & 68.9                     & 14.60                     \\
			vegetation-\textgreater{}road              & 59.14                   & 74.32                  & 75.55                    & 73.13                    & 60.10                    & 75.08                  & 78.81                    & 71.68                    \\
			building-\textgreater{}road                & 42.10                    & 59.25                  & 81.97                    & 46.40                     & 42.27                   & 59.42                  & 76.50                     & 48.57                    \\
			farmland-\textgreater{}road                & 69.94                   & 82.31                  & 86.16                    & 78.79                    & 71.04                   & 83.07                  & 83.96                    & 82.2                     \\
			grass-\textgreater{}road                   & 28.98                   & 44.93                  & 81.75                    & 30.98                    & 40.78                   & 57.93                  & 81.93                    & 44.81                    \\
			road-\textgreater{}grass                   & 22.78                   & 37.11                  & 83.44                    & 23.86                    & 21.87                   & 35.89                  & 84.47                    & 22.79                    \\
			bridge-\textgreater{}water                 & 75.10                    & 85.78                  & 89.44                    & 82.41                    & 80.01                   & 88.90                   & 89.56                    & 88.24                    \\
			\midrule
			mIoU                          & \multicolumn{4}{c|}{55.12}                                                                              & \multicolumn{4}{c}{55.71}  \\
			\bottomrule
		\end{tabular}
	}
	\label{tab:LWGANet}
\end{table}
\begin{table}[t!]
	\centering
	\scriptsize
	\caption{Comparison of quantitative performance for binary change detection         
		on RB-SCD.
	}
	\begin{tabular}{c|cccc}
		\toprule
		Model        & IoU   & F1    & Rec   & Pre   \\
		\midrule
		SNUNet~\cite{SNUNet}          & 62.90 &77.22 & 72.47 & 82.64 \\
		BIT~\cite{BIT}          & 62.21 & 76.71 & 70.97 & 83.47 \\
		ChangeFormer~\cite{Changeformer} & 61.88 & 76.45 & 69.69 & 84.66 \\
		ICIF~\cite{ICIF}         & 61.80  & 76.39 & 71.93 & 81.43 \\
		TFI-GR~\cite{TFI-GR}       & 64.72 & 78.58 & 73.78 & 84.05 \\
		DMINet~\cite{DMINet}       & 62.99 & 77.29 & 72.14 & 83.24 \\
		USSFC-Net~\cite{USSFC-Net}    & 62.55 & 76.96 & {\color{red}75.07} & 78.95 \\ 
		ELGC-Net~\cite{ELGC-Net}     & {\color{blue}65.53} & {\color{blue}79.18}& 73.99 & {\color{blue}85.15} \\
		\midrule
		Ours        &{\color{red}65.83 }& {\color{red}79.39} & {\color{blue}74.13} & {\color{red}85.47}\\
		\bottomrule
	\end{tabular}
	\label{tab:BCD}
\end{table}
\begin{table*}[t]
	\caption{Comparison of quantitative performance of various change detection techniques on public datasets. 
		Results marked with * are obtained from the original papers. The ‘-' symbol indicates that the corresponding data is not available in the source. }
	
	\centering
		\resizebox{1\textwidth}{!}{
			\begin{tabular}{c|c|c|c}
				\toprule
				\multirow{2}{*}{Model}   & LEVIR-CD                            & SYSU-CD                            & WHU-CD                              \\
				& IoU\quad\quad F1\quad\quad Rec\quad\quad Pre                & IoU\quad\quad F1\quad\quad Rec\quad\quad Pre                & IoU\quad\quad F1\quad\quad Rec\quad\quad Pre                 \\
				\midrule
				SNUNet~\cite{SNUNet}                  & 82.20\quad90.23\quad88.97\quad91.54          & 66.73\quad80.04\quad79.79\quad80.30          & 76.02\quad86.38\quad88.95\quad83.95          \\
				BIT~\cite{BIT}                     & 82.06\quad90.15\quad88.98\quad91.34          & 62.50\quad76.93\quad72.50\quad81.93          & 80.40\quad89.13\quad89.69\quad88.58          \\
				ChangeFormer~\cite{Changeformer}           & 81.93\quad90.07\quad87.38\quad{\color{red}92.92}          & 60.46\quad75.36\quad71.06\quad80.22          & 83.88\quad91.23\quad90.76\quad91.71          \\
				
				ICIF~\cite{ICIF}                     & 79.86\quad88.80\quad86.85\quad90.84          & 61.38\quad76.07\quad73.31\quad79.05          & 78.27\quad87.81\quad84.06\quad91.96          \\
				TFI-GR~\cite{TFI-GR}                   & 83.03\quad90.73\quad89.53\quad91.95          & {\color{blue}71.38\quad83.30}\quad81.16\quad{\color{blue}85.56}          & {\color{blue}88.55\quad93.92}\quad92.65\quad{\color{red}95.24}         \\
				DMINet~\cite{DMINet}                   & 83.14\quad90.79\quad88.89\quad{\color{blue}92.78}          & 68.20\quad81.10\quad76.82\quad{\color{red}85.88}         & 79.53\quad88.60\quad91.24\quad86.10          \\
				USSFC-Net~\cite{USSFC-Net}                   & 82.54\quad90.44\quad91.82\quad89.09          &65.75\quad79.34\quad{\color{blue}81.52}\quad77.27         & 83.92\quad91.26\quad93.20\quad89.40         \\
				
				ELGC-Net~\cite{ELGC-Net}                    & 83.52\quad91.02\quad89.78\quad92.30          &65.34\quad79.03\quad77.26\quad80.89         & 75.56\quad86.08\quad82.15\quad90.40          \\
				MDENet *~\cite{MDENet} & {\color{blue}83.63\quad91.09}\quad{\color{red}92.68}\quad89.55  &-\quad\quad\quad-\quad\quad\quad-\quad\quad\quad-
				&87.74\quad93.47\quad{\color{blue}95.49}\quad91.54\\
				MHF²Net *~\cite{MHFNet}                   & 82.64\quad90.50\quad\thickspace\thickspace\thickspace-\quad\quad\thickspace\thickspace\thickspace\thickspace-\thickspace\medspace\medspace\medspace        &-\quad\quad\quad-\quad\quad\quad-\quad\quad\quad-        & 87.14\quad93.13\quad\thickspace\thickspace\thickspace-\quad\quad\thickspace\thickspace\thickspace\thickspace-\thickspace\medspace\medspace\medspace              \\
				
				\midrule
				Ours                   & {\color{red}84.69\quad91.71}\quad{\color{blue}92.31}\quad91.11 & {\color{red}72.63\quad84.15\quad85.60\quad}82.74 & {\color{red}90.43\quad94.97\quad96.44\quad}{\color{blue}93.55}\\
				\bottomrule
		\end{tabular}}
	
	\label{tab:public datasets}
\end{table*}
68.40\%, respectively, significantly outperforming other comparison methods. These results highlight the strong capability of our model in semantic discrimination and spatial reasoning.
More importantly, our method also performs excellently on several challenging categories, including \textit{road-\textgreater{}vegetation}, \textit{road-\textgreater{}grass}, \textit{road-\textgreater{}bare land}, and \textit{grass-\textgreater{}road}, which suffer from sample scarcity and semantic ambiguity. These types of changes are often affected by similar surface textures, seasonal variability, and complex contextual information. Traditional methods are prone to false positives and missed detections under such conditions, whereas our model consistently demonstrates strong robustness and generalization ability by accurately identifying these subtle semantic transformations.
Furthermore, we evaluate the performance of the latest RS foundation model, LWGANet~\cite{LWGANet}, on our proposed dataset (as shown in Tab.~\ref{tab:LWGANet}). The results further underscore the challenging nature of our dataset. 

\textbf{Results for Binary Change Detection on RB-SCD Dataset.}
As shown in Tab.~\ref{tab:BCD}, in the BCD task based on the RB-SCD dataset, different models exhibit varying performance across different evaluation metrics. In terms of the IoU metric, our model ranks first with a score of 65.83\%, slightly outperforming ELGC-Net~\cite{ELGC-Net}, which achieves 65.53\%. This indicates that MFDCD demonstrates the best performance in terms of overlap between the predicted and ground truth change regions, enabling more precise localization of change areas.
Regarding the F1, our MFDCD model also leads with a score of 79.39\%, followed closely by ELGC-Net at 79.18\%. Since the F1 is a harmonic mean of Pre and Rec, this result suggests that MFDCD achieves an excellent balance between detection accuracy and completeness.

\textbf{Results on Public CD Datasets.} To further prove the validity of our model, we also compare our proposed MFDCD with ten state-of-the-art CD models. To ensure a fair comparison, we reproduce these methods using publicly available code with default parameters.  
As observed in Tab.~\ref{tab:public datasets}, our method achieves outstanding results across all three publicly available datasets. Specifically, on the WHU-CD dataset, our approach outperforms the second-best method by 2.69\% and 1.5\% in IoU and F1, respectively. 
On the SYSU-CD dataset, it achieves improvements of 4.43\% and 3.05\%, while on the LEVIR-CD dataset, it surpasses the second-best method by 1.06\% and 0.62\%. 
These results further demonstrate the effectiveness and robustness of our method.
\begin{figure*}[t]
	\centering
	\includegraphics[width=1\linewidth]{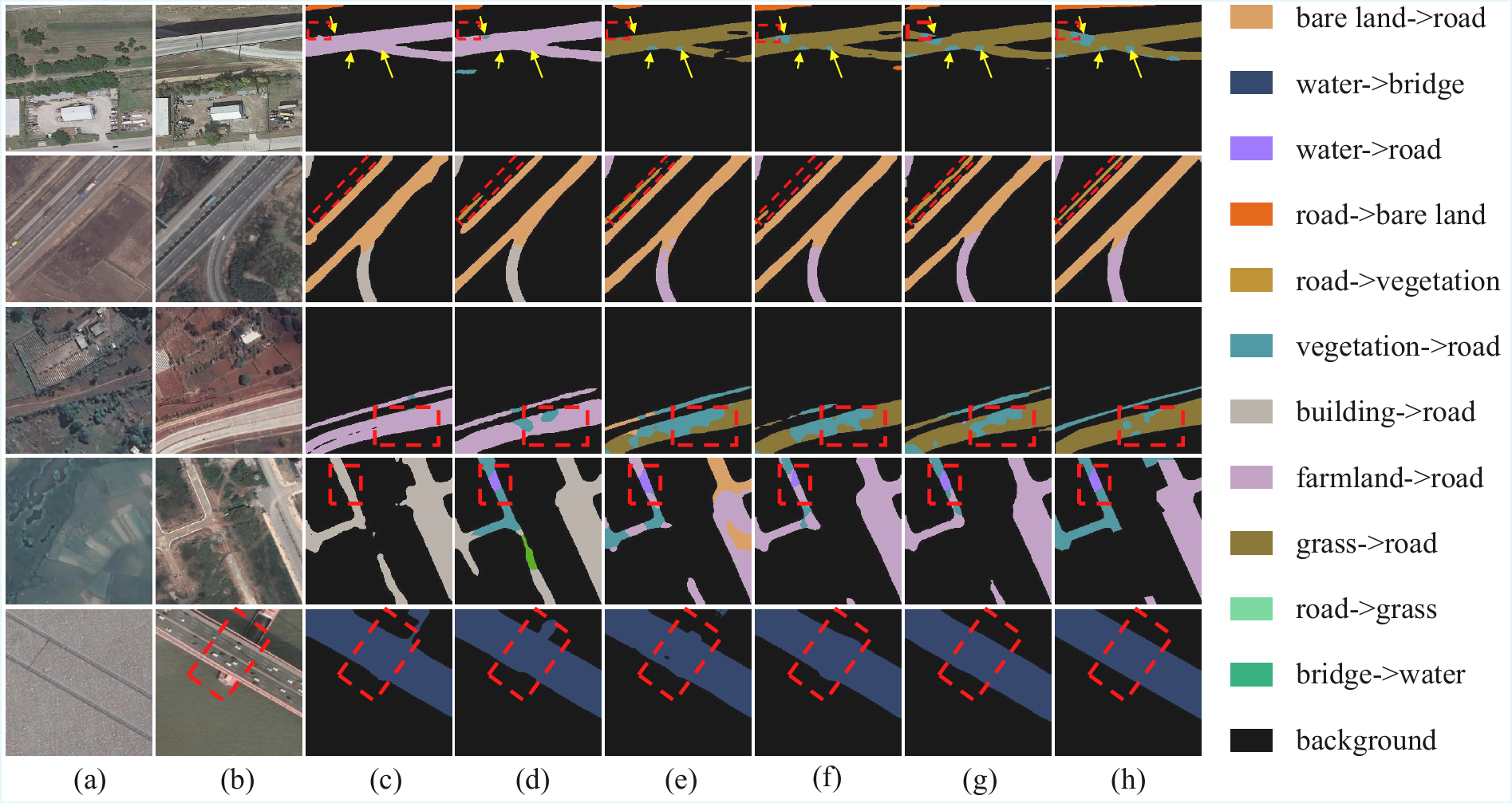} 
	\caption{Qualitative visualization for semantic change detection on RB-SCD. (a) $T_1$ images. (b) $T_2$ images. 
		(c) HGINet. (d) CdSC. (e) DEFO. (f) ChangeCLIP. (g) Ours. (h) Label.}
	\label{fig:srbsd_visual_results}
\end{figure*}

\subsection{Qualitative Analysis.} 
\textbf{Qualitative Visualization for BCD Results.} Fig.~\ref{fig:srbsd_visual_results} presents a qualitative comparison of MFDCD against  state-of-the-art methods on challenging examples from the RB-SCD dataset. The visualizations highlight the tangible benefits of our multimodal, frequency-driven design.

Superior Delineation of Linear Transitions: The first two rows demonstrate the effectiveness of our DFC in modeling linear and heterogeneous change transitions. In the first row, where a new road cuts through grass, other methods produce fragmented or incomplete predictions, especially under the road's shadow. In contrast, our MFDCD, empowered by the DFC's ability to model the long-range  continuity of linear structures via its BTFF block, generates a much more complete and coherent road segment. Similarly, in the second row (bare land-\textgreater{}road), the DFC's ASFF block, which is sensitive to high-frequency boundary details, allows for a sharper and more precise delineation of the road edges compared to other methods.

Robustness to Semantic Ambiguity: The third and fourth rows showcase the critical role of our TFF in resolving high semantic ambiguity. The third-row case (vegetation-\textgreater{}road and grass-\textgreater{}road) is a classic example of a scene where multiple land covers transition into a road. While other methods produce a confusing mix of class predictions, our TFF, acting as a semantic reasoning engine, provides a strong contextual prior. This guides the model to a confident and correct interpretation of the scene, resulting in a cleaner and more accurate segmentation. The fourth row further illustrates this, our method correctly identifies the subtle water-\textgreater{}road change (light green) that other methods, including the language-guided ChangeCLIP, largely miss, demonstrating the effectiveness of our frequency-domain semantic alignment.

Handling Pseudo-Changes and Complex Scenes: The final row (water-\textgreater{}bridge) highlights the  robustness of our model. The clean, sharp prediction of the new bridge demonstrates the DFC's ability to extract the essential structural form while being less sensitive to noise like water texture.

\begin{figure*}[t]
	\centering
	\includegraphics[width=1\linewidth]{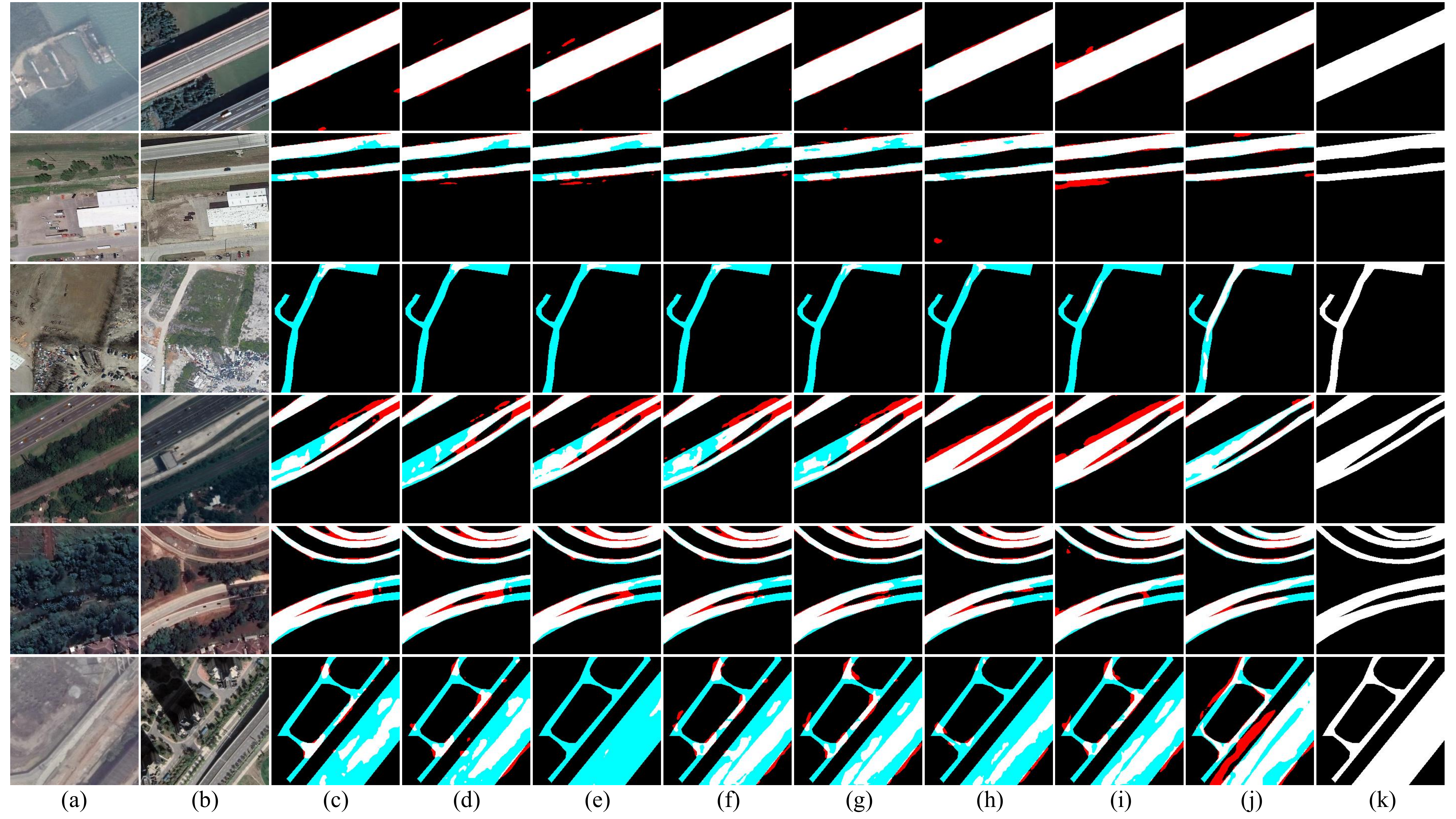} 
	\caption{Qualitative visualization for binary change detection on RB-SCD dataset. (a) $T_1$ images. (b) $T_2$ images. 
		(c) BIT.  (d) ChangeFormer. (e) ICIF. (f) TFI-GR. (g) DMINet. (h) USSFC-Net. (i) ELGC-Net. (j) Ours. (k) Label. The rendered colors represent true positives (white), false positives (red), true negatives (black), and false negatives (blue).}
	\label{fig:bcd_visual_results}
	
\end{figure*}

\textbf{Qualitative Visualization for BCD Results.} As shown in Fig.~\ref{fig:bcd_visual_results}, it can be observed that 
 in areas where buildings are newly constructed or modified, most models can capture at least part of the change, though the precision of boundary detection varies.
In regions where there is a transition from bare land to road, our MFDCD and some others show better performance in accurately delineating the change area. Some methods, however, have varying degrees of missed detections. For instance, methods such as BIT~\cite{BIT} and ChangeFormer~\cite{Changeformer} seem to have less accurate delineation compared to our method and ELGC-Net~\cite{ELGC-Net}.
Some models, like ICIF~\cite{ICIF} and TFI-GR~\cite{TFI-GR}, can detect the change but with somewhat blurred boundaries compared to the ground truth (white regions representing true positives). Other models may miss parts of the change.
In areas where the change from bare land to road is affected by shadows, most models struggle. Except for our MFDCD, which shows relatively better performance in distinguishing true changes from pseudo-changes caused by shadows, other models either misclassify the shadow areas as changes (false positives in red) or miss the actual changes (false negatives in blue).

\begin{table}[t]
	\caption{ Ablation studies of DFC and TFF  when trained on RB-SCD and  public
		CD datasets. 
	}
	\centering
		\scalebox{0.95}{
	\begin{tabular}{ccccccc}
		\toprule
		\multirow{1}{*}{Method} & \multirow{1}{*}{DFC} & \multirow{1}{*}{TFF} & \multirow{1}{*}{RB-SCD} & \multirow{1}{*}{LEVIR-CD} & \multirow{1}{*}{SYSU-CD} & \multirow{1}{*}{WHU-CD} \\
	                                                               
		\midrule
		(a)& \ding{55}& \ding{55} & 50.45& 81.65 & 68.98  & 86.65                                                              \\

		(b)& \checkmark& \ding{55}  & {\color{blue}55.89} & {\color{blue}83.25}  & {\color{blue}71.10}  & {\color{blue}89.28}                                                              \\
		
		(c) & \ding{55}  & \checkmark  & 55.23  & 82.97 & 70.89  & 88.50  
		                                                             \\

		(d)&\checkmark& \checkmark & {\color{red}57.52}& {\color{red}84.69}& {\color{red}72.63}& {\color{red}90.43}                                         \\
		\bottomrule
	\end{tabular}}
	
	\label{tab:results_table_4}
\end{table}
\begin{table}[t]
	\caption{Effectiveness of the DFC module when trained on RB-SCD and public CD datasets. 
	}
	\centering
	\resizebox{1\textwidth}{!}{
		\begin{tabular}{cccccc}
			\toprule
			\multicolumn{2}{c}{\multirow{1}{*}{Method}} & \multirow{1}{*}{RB-SCD} & \multirow{1}{*}{LEVIR-CD} & \multirow{1}{*}{SYSU-CD} & \multirow{1}{*}{WHU-CD} \\
			\midrule
			(a)               & -                       & 55.23                   & 82.97                                                                & 70.89                                                               & 88.50                                                               \\
			(b)               & $\mathbf{F}_i^1+\mathbf{H_i^2}, i \in \left \{ 1,2,3,4 \right \} $                        & 55.84                   & 83.52                                                                & 71.42                                                               & 89.34                                                              \\
			(c)               & $\mathbf{F}_i^1+\mathbf{H_i^2}, \mathbf{F}_i^2+\mathbf{H_i^1}, i \in \left \{ 1,2\right \}  $                        & 55.70                   & 83.39                                                                & 71.23                                                               & 88.90                                                             \\
			(d)               & $\mathbf{F}_i^t+\mathbf{LL}_i^t, i \in \left \{3,4 \right \} $                        & 55.86                    & 83.51                                                                & 71.30                                                               & 89.24                                                              \\
			(e)               & ASFF                    & 56.42                   & {\color{blue}84.02}                                                                & {\color{blue}72.06}                                                               & {\color{blue}89.98}                                                              \\

			(f)               & BTFF                    & {\color{blue}56.87}                   & 83.90                                                                 & 71.85                                                               & 89.77                                                              \\
			(g)& TFIM +TFF& 55.98 & 83.86  & 71.63 & 88.97                                                              \\
			(h)               & ASFF+BTFF (DFC)             & {\color{red}57.52}& {\color{red}84.69}& {\color{red}72.63}& {\color{red}90.43}      \\
			\bottomrule
	\end{tabular}
}
	\label{tab:tab8}
\end{table}

\begin{table}[t]
	\caption{Effectiveness of the TFF module when trained on RB-SCD and public CD datasets. 
	}
	\centering
	\resizebox{1\textwidth}{!}{
		\begin{tabular}{cccccc}
			\toprule
			\multicolumn{2}{c}{\multirow{1}{*}{Method}} & \multirow{1}{*}{RB-SCD} & \multirow{1}{*}{LEVIR-CD} & \multirow{1}{*}{SYSU-CD} & \multirow{1}{*}{WHU-CD} \\
			\midrule
					(a)                     & -                  & 55.89                   & 83.25                                                                & 71.10                                                                & 89.28                                                              \\
(b)& DFT+MLP  &56.31  &83.59  &71.42 &89.57                                                            \\
(c)& DFT+1D CNN  &56.47  &83.73  &71.62   & 89.68                                                       \\
(d)& DFT+GAT   &{\color{blue}56.67}  &{\color{blue}83.93}  &71.69   & 89.75                                                       \\
(e)& CLIP Embedding + Concat  & 56.38 & 83.54 &  {\color{blue}71.85} & {\color{blue}89.84}                                                             \\
		(f)& DFT + Graph Filter (TFF) & {\color{red}57.52}& {\color{red}84.69}& {\color{red}72.63}& {\color{red}90.43}                                         \\
			\bottomrule
	\end{tabular}}
	\label{tab:tab9}
\end{table}

\begin{table}[t]
	\caption{Effectiveness of graph  filters with different numbers $l$ in TFF.
	}
	\centering
	\scalebox{0.9}{
	\begin{tabular}{cccccc}
		\toprule
		Method & $l$ & RB-SCD & LEVIR-CD & SYSU-CD & WHU-CD \\     
		\midrule
		(a)                     & -                  & 55.89                   & 83.25                                                                & 71.10                                                                & 89.28                                                              \\
		(b)                     & 3                  & 56.74                   & 83.84                                                                & 71.98                                                               & 89.87                                                              \\
		(c)                     & 4                  & 57.33                   & 84.13                                                                & {\color{blue}72.31}                                                               & 90.10                                                              \\
		(d)                     & 5                 &{\color{red}57.52}& {\color{red}84.69}& {\color{red}72.63}& {\color{red}90.43}                                                            \\
		(e)                     & 6                  & {\color{blue}57.40}                    & {\color{blue}84.26}                                                                & 72.25                                                               & {\color{blue}90.38}      \\
		\bottomrule
 	\end{tabular}}
	\label{tab:results_table_6}
\end{table}
\begin{table}[t]
	\centering
	\caption{
		Effectiveness of MFDCD with different backbones when trained on RB-SCD and  public CD datasets.
	}
			\resizebox{1\textwidth}{!}{
	\begin{tabular}{ccccccc}
		\toprule
		\multirow{1}{*}{Backbone} &Params (M) & FLOPs (G)& RB-SCD & LEVIR-CD & SYSU-CD & WHU-CD \\
		\midrule
		ResNet18& 13.75 & 11.32& 55.53 & 83.18 & 71.35 & 89.04\\
		ResNet34 & 22.80 &16.54& {\color{blue}56.31}&{\color{blue} 83.78}& {\color{blue}72.10}  & {\color{blue}89.73}  \\ResNet50  & 103.78       &  49.02 &  {\color{red}57.52 } &  {\color{red} 84.69}&  {\color{red} 72.63 } &  {\color{red} 90.43 } \\
		\bottomrule
	\end{tabular}}
	\label{tab:backbone}
\end{table}
\subsection{Ablation Study}To demonstrate the effectiveness of our proposed modules, we conduct extensive experiments on RB-SCD and three public datasets. 

\textbf{Effectiveness of DFC and TFF Modules.} As shown in Tab.~\ref{tab:results_table_4}, adding the DFC and TFF  to the baseline model increases the mIoU on the RB-SCD dataset by  5.44\% and 4.78\%, respectively. Additionally, the IoU on the LEVIR-CD, SYSU-CD, and WHU-CD datasets improved by 1.6\%/2.12\%/2.63\% and 1.32\%/1.91\%/1.85\%, respectively. 

\textbf{Detailed Analysis on DFC  Module.} As shown in Tab.~\ref{tab:tab8}, we conduct detailed ablation studies on the ASFF and BTFF blocks in the DFC module. Initially, we experiment with different coupling methods for ordinary features and frequency domain features. By comparing rows (b) and (c) with row (h), we can observe that ASFF and BTFF effectively dynamically couple visual features with frequency domain features, achieving the best results. Comparing rows (c) and (e), it can be seen that combining the low-level features at time $T_1$ with the corresponding high-frequency features at time $T_2$, due to semantic mismatches, results in worse performance compared to our designed ASFF. Comparing rows (d) and (f), it is clear that our designed BTFF captures the global structural changes across time, outperforming the coupling of high-level features and low-frequency features at the same time point.
To further validate our design, we replace our DFC with TFIM~\cite{TFI-GR} (row (g)), a strong module for spatial feature interaction. Although TFIM provides a notable improvement over the baseline, our frequency-domain DFC  yields  better performance, demonstrating the superiority of explicitly modeling structural features in the frequency domain for this task.

\textbf{Detailed Analysis on TFF Module.} As shown in Tab.~\ref{tab:tab9}, the MLP (row (b)), which processes each frequency component independently, performs the worst, highlighting the necessity of modeling relational information. The 1D CNN (row (c) captures only local relationships and offers a slight improvement.
Notably, while the GAT adaptively learns edge weights (row (d)), our proposed method (row (e)) with a simple graph filter over a fixed fully connected graph still achieves the best results. 
This indicates long-range dependency across the entire spectrum is more effective than attempting to dynamically prune or re-weight connections
To validate our frequency-domain approach, we benchmark our TFF module against a powerful CLIP Embedding + Concat~\cite{ICIF} mechanism (row (e)) that fuses semantics in the spatial/token domain. It shows that while this mechanism improves upon the vision-only baseline, our TFF-based model (row (f)) achieves superior performance. This suggests that extracting a robust, global semantic signature via frequency analysis is more effective for disambiguating complex RS scenes than performing direct spatial alignment with text tokens. As shown in 
\begin{figure*}[t]
	\centering
	\includegraphics[width=1\linewidth]{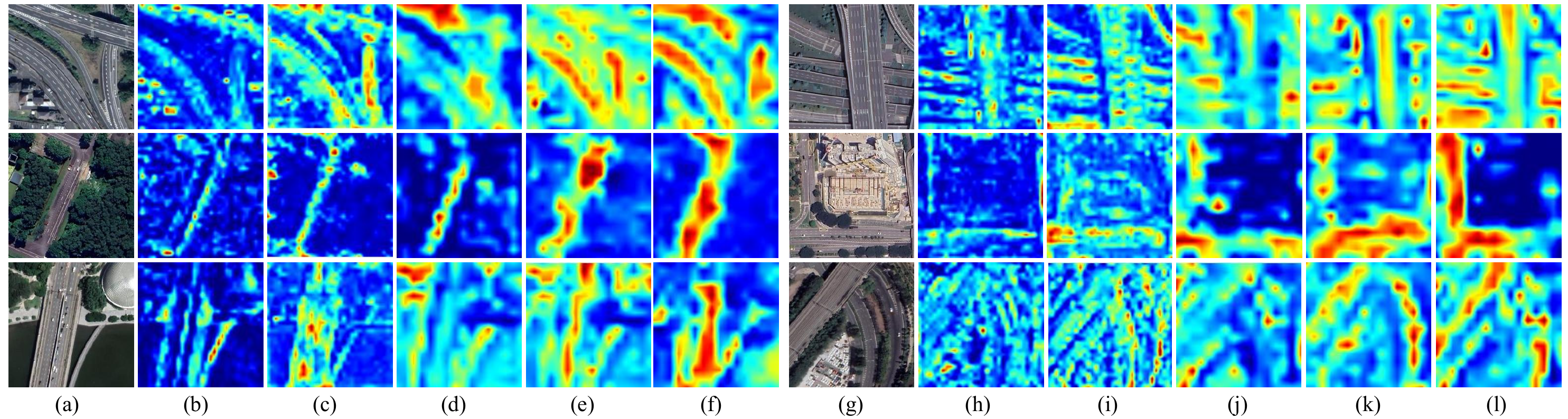} 
	\caption{Heatmaps of our proposed modules. (a) and (g) Input images.  (b) and (h) Low-level features from the backbone. (c) and (i)  High-level features from the backbone network. (d) and (j) Feature maps after undergoing the ASFF block. (e) and (k) Feature maps after undergoing the BTFF block. (f) and (l)  Feature maps after undergoing the TFF module.}
	\label{fig:heatmap}
\end{figure*}
\begin{figure*}[t]
	\centering
	\includegraphics[width=1\linewidth]{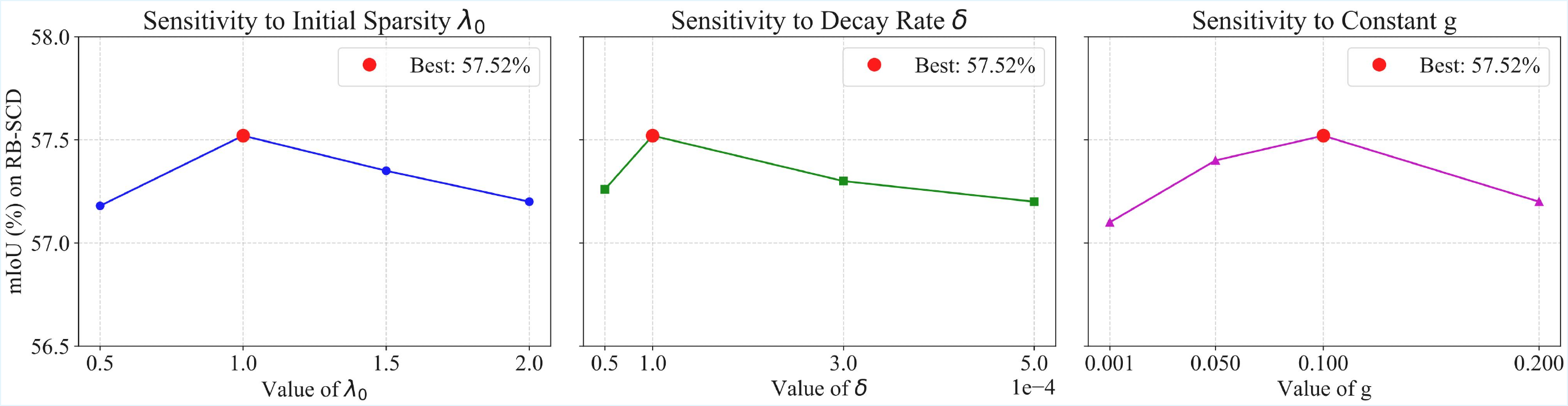} 
	\caption{Hyperparameter sensitivity analysis of the dynamic sparsity constraint in ASFF. We evaluate the impact of the initial sparsity $\lambda_0$ (left), the decay rate $\delta$ (middle) and the constant $g$ (right) on the mIoU performance on  RB-SCD.}
	\label{fig:DFC_Params}
\end{figure*}
Tab.~\ref{tab:results_table_6}, we conduct ablation experiments on the number of layers in the frequency filter of TFF. As the value of $l$ increased, the model's performance improved. However, when the number of filter layers reached six, the performance stabilized or slightly declined. Subsequent experiments are based on the optimal setting, \( l=5 \).
 
\textbf{Backbone Analysis.}
To investigate the impact of backbone networks  of the MFDCD model, as shown in Tab.~\ref{tab:backbone}, in terms of parameters and computational complexity, ResNet18 has the lowest cost, while ResNet50 has the highest. In terms of detection accuracy, 
for all four datasets, 
the deeper the backbone network, the higher the accuracy. However, a deeper network also leads to a doubling of parameters and computational complexity. In this paper, 
ResNet50 is selected as the backbone network.

\textbf{Visualization Analysis.} To more clearly illustrate the influence of our proposed modules, we visualized heatmaps for these modules. As depicted in columns (b)/(h) and (c)/(i) of Fig.~\ref {fig:heatmap}, the low-level and high-level features from the backbone network typically exhibit certain characteristics. The low-level features may contain fine-grained but scattered information, while the high-level features, though more abstract, might lose some detailed spatial information. In columns (d)/(j), (e)/(k), and (f)/(l), after undergoing the ASFF block, BTTF block and TFF module respectively, the feature maps show progressive changes. The ASFF block helps to fuse features in a reasonable way, precisely delineate the boundaries. The BTTF block further maintains the continuity and integrity of linear structures. The TFF module finally outputs feature maps that can better highlight the change regions, eliminating some irrelevant noise and resolving semantic ambiguities.


\textbf{Hyperparameter  Analysis.} 
As shown in Fig.~\ref{fig:DFC_Params}, we conduct a sensitivity analysis on the key hyperparameters $\lambda_0$, $\delta$ and $g$ of our dynamic sparsity constraint. The performance remains stable across a reasonable range of values, demonstrating the robustness of our approach. Our chosen values ($\lambda_0$=1.0, $\delta$=0.0001,$g$=0.1) provide the best performance. This dynamic mechanism proves more effective than a fixed threshold, as it adapts the level of detail refinement to the model's learning stage, ensuring both stability in early training and precision in later stages.

\section{Conclusion}
In this work, we introduce RB-SCD, a SCD dataset specifically designed for complex traffic scenarios. RB-SCD provides comprehensive fine-grained pixel-level annotations across 11 semantic change categories, encompassing a wide range of road and bridge types. It also covers diverse geographic regions across multiple countries and urban environments, making it a challenging  benchmark for road and bridge change analysis in traffic scenes. Notably, RB-SCD supports both BCD and SCD tasks, offering flexibility for different application scenarios. 
We believe this benchmark represents an important first step towards advancing SCD of roads and bridges, and we envision future extensions to larger-scale and more balanced datasets.  

In addition, we propose the MFDCD framework, which pioneers a new paradigm by integrating multimodal features in the frequency domain. It tackles the core challenges with two specialized components: the DFC module, which leverages wavelet analysis to robustly model the evolution of linear structures, and the TFF module, which acts as a semantic reasoning engine to disambiguate visually similar land covers. Extensive experiments on RB-SCD and three public datasets validate the superiority of MFDCD over state-of-the-art methods, highlighting its robustness, generalization, and strong discrimination power.  

\textbf{Limitation and Feature Work. }Nevertheless, we acknowledge its limitations. The performance of MFDCD still degrades under extreme shadow or for rare, under-represented change classes, and the framework is not yet lightweight enough for deployment in resource-constrained scenarios. 
Future work will therefore focus on three directions: (1) expanding RB-SCD with more rare-class samples and temporal sequences; (2) extending our frequency-driven paradigm to topology-aware change reasoning for better connectivity analysis and pseudo-intersection recognition; and (3) developing lightweight model variants via compression and efficient architecture design to enhance applicability in real-time and edge-computing scenarios.

\section*{Acknowledgement}
This work was supported in part by the NSFC Key Project of Joint Fund for Enterprise Innovation and Development under Grant U24A20342, and in part by the National Natural Science Foundation of China under Grant 62576006 and 61976004.

\bibliography{mybibfile}

\end{document}